\documentclass[journal]{IEEEtran}
\usepackage{multirow}
\usepackage{times}
\usepackage{epsfig}
\usepackage{graphicx}
\usepackage{amsmath}
\usepackage{amssymb}
\usepackage{verbatim}
\usepackage{subcaption}
\usepackage{bbding} 
\usepackage{pifont}
\usepackage{wasysym}
\usepackage{bm}
\usepackage{float}
\usepackage{multirow}
\usepackage[export]{adjustbox}
\usepackage{soul}
\usepackage{csquotes}
\usepackage{xspace}

\newcommand{\latinphrase}[1]{\textit{#1}}

\newcommand{\ie}{\latinphrase{i.e.,}\xspace}

\newcommand{\etc}{\latinphrase{etc.}\xspace}

\usepackage[colorlinks,linkcolor={},pagebackref=true,breaklinks=true,colorlinks,bookmarks=false]{hyperref}

\newcommand{\z}{\mathbf{z}}
\newcommand{\zc}{\mathbf{z}_C}
\newcommand{\zn}{\mathbf{z}_N}
\newcommand{\hatz}{\mathbf{\hat{z}}}
\newcommand{\y}{\mathbf{y}}
\newcommand{\x}{\mathbf{x}}
\newcommand{\n}{\mathbf{n}}
\newcommand{\sa}{\mathbf{s}_1}
\newcommand{\biasa}{\mathbf{b}_1}

\newcommand{\hi}{\mathbf{h}_i}
\newcommand{\hii}{\mathbf{h}_{i+1}}
\newcommand{\ngaussian}{\mathcal{N}(\mathbf{0}, \mathbf{I})}

\def\ourmodel{FDN}

\ifCLASSINFOpdf
\else
\fi

\begin{document}
\title{Disentangling Noise from Images: \\
A Flow-Based Image Denoising Neural Network}

\author{Yang~Liu,
        Saeed~Anwar,
        Zhenyue~Qin, 
        Pan Ji, 
        Sabrina Caldwell,
        and~Tom~Gedeon
\thanks{Y. Liu, S. Anwar, Z. Qin, S. Caldwell and T. Gedeon are with the Research School of Computer Science, the Australian National University, Canberra, ACT 2600, Australia.}
\thanks{Y. Liu and S. Anwar are also with Data61-CSIRO, Australia}
\thanks{P. Ji is with OPPO US Research, America}
\thanks{Corresponding Author: yang.liu3@anu.edu.au}}

\maketitle


\begin{abstract}

The prevalent convolutional neural network (CNN) based image denoising methods extract features of images to restore the clean ground truth, achieving high denoising accuracy. However, these methods may ignore the underlying distribution of clean images, inducing distortions or artifacts in denoising results. This paper proposes a new perspective to treat image denoising as a distribution learning and disentangling task. Since the noisy image distribution can be viewed as a joint distribution of clean images and noise, 
the denoised images can be obtained via manipulating the latent representations to the clean counterpart. 
This paper also provides a distribution learning based denoising framework. Following this framework, we present an invertible denoising network, \ourmodel{}, without any assumptions on either clean or noise distributions, 
as well as a distribution disentanglement method. \ourmodel{} learns the distribution of noisy images, which is different from the previous CNN based discriminative mapping. 
Experimental results demonstrate \ourmodel's capacity to remove synthetic additive white Gaussian noise (AWGN) on both category-specific and remote sensing images. Furthermore, the performance of \ourmodel{} surpasses that of previously published methods in real image denoising with fewer parameters and faster speed. Our code is available at: 
\href{https://github.com/Yang-Liu1082/FDN.git}{https://github.com/Yang-Liu1082/FDN.git}.
\end{abstract}

\begin{IEEEkeywords}
Image Denoising, Invertible Network, Normalizing Flow.
\end{IEEEkeywords}

\IEEEpeerreviewmaketitle


\section{Introduction}
\IEEEPARstart{D}{espite} decades of research, image denoising~\cite{Anwar2019RIDNET,liu2020gradnet} is still an on-going low-level image processing task in computer vision. The long-standing interest in image denoising has provided roots for a vast array of downstream applications, such as segmentation~\cite{minaee2020image} and deblurring~\cite{Zhang_2017_CVPR}. Nearly all images need to be denoised before further processing, especially those obtained in dark environments. 

\begin{figure}[ht]
\centering
\begin{subfigure}{.155\textwidth}
  \centering
  \includegraphics[width=\linewidth]{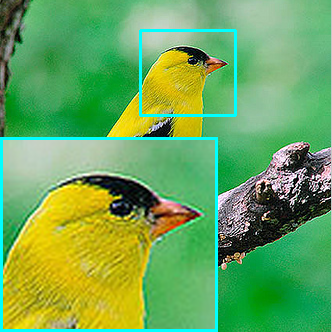}
  \caption{Ground-truth}
  \vspace{2mm}
\end{subfigure}
\begin{subfigure}{.155\textwidth}
  \centering
  \includegraphics[width=\linewidth]{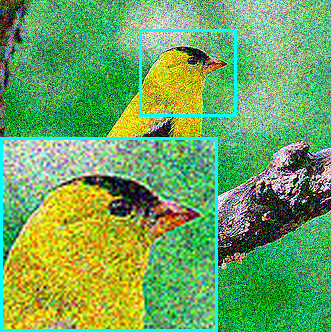}
  \caption{Noisy}
  \vspace{2mm}
\end{subfigure}
\begin{subfigure}{.155\textwidth}
  \centering
  \includegraphics[width=\linewidth]{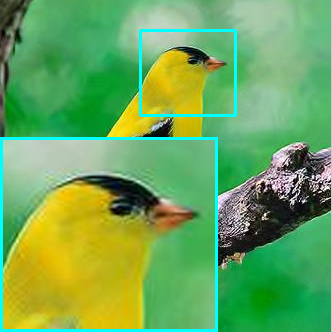}
  \caption{BM3D~\cite{BM3D}}
  \vspace{2mm}
\end{subfigure}
\begin{subfigure}{.155\textwidth}
  \centering
  \includegraphics[width=\linewidth]{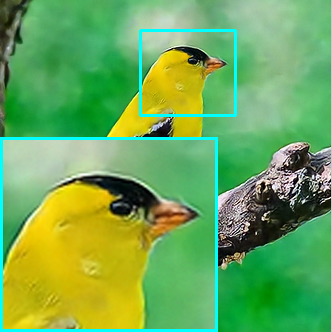}
  \caption{DnCNN~\cite{DnCNN}}
  \vspace{2mm}
\end{subfigure}
\begin{subfigure}{.155\textwidth}
  \centering
  \includegraphics[width=\linewidth]{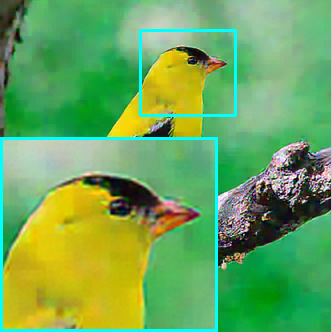}
  \caption{IRCNN~\cite{IRCNN}}
  \vspace{2mm}
\end{subfigure}
\begin{subfigure}{.155\textwidth}
  \centering
  \includegraphics[width=\linewidth]{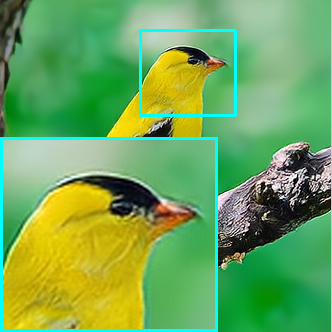}
  \caption{\ourmodel{} (Ours)}
  \vspace{2mm}
\end{subfigure}
\caption{The visual comparison on CUB-200~\cite{Bird} with $\sigma=50$ AWGN. 
Our method restores finer feathers, clearer eyes and the sharper beak. Zooming in on a high-resolution display will allow better observation of the differences.
}
\vspace{-6mm}
\label{fig:Visual_Front}
\end{figure}

The purpose of image denoising is to reconstruct clean images from corrupted noisy observations.
Traditional denoising methods rely on certain assumptions on noise distributions~\cite{6751276,7557070} or priors on ground truth clean images~\cite{BM3D,6122031} to build optimization models. However, these assumptions and priors may differ from the real case, which can compromise the denoising accuracy.
Deep learning denoising approaches proposed in recent years use convolutional neural networks (CNNs) to learn the models from a large number of noise-free and noisy image pairs, and have achieved superior denoising performance~\cite{DnCNN,zhang2018ffdnet}.
These methods employ CNNs to learn the mapping functions between noisy images and clean ones. However, they usually overemphasize the pixel similarity between the denoised image and the clean ground truth while omitting the underlying distribution of clean images. Thus, although some deep methods can obtain high quantitative results, over-smoothed regions and artifacts are often brought into the restored images, resulting in degraded visual results.
 
This paper reconsiders image denoising from the perspective of distribution disentanglement.
The distribution of noisy images can be treated as a joint distribution of clean images and noise. Thus, it is intuitive to consider conducting image denoising via disentangling these two distributions. 
Following this line of thought, the process of distribution-learning based image denoising can be divided into three stages: the first is to learn the distribution of noisy images by transforming the noisy images into latent representations; the second is to disentangle the representation of clean images from the noisy ones; and the last is to restore clean images from the disentangled clean representation. 

There are two challenges for us to overcome: which kind of network is suitable for learning the distributions and restoring images, and how to disentangle the two distributions.
For the first problem, we resort to generative models to learn the distributions. We require the generative model to generate a denoised image given a disentangled latent code. The denoised image should follow a clean image distribution and be visually similar to the corresponding noisy image. Therefore, the candidate generative model should have a one-to-one mapping between the noisy image space and the latent space. Also, a subspace of the latent space, \ie the space of the disentangled latent code for clean images, can also be one-to-one mapped to the clean image space. A variational autoencoder (VAE)~\cite{VAE} cannot guarantee the one-to-one mapping between the latent representations and images. On the other hand, although generative adversarial network (GAN) can ensure the one-to-one mapping~\cite{ma2018invertibility}, learning two different distributions, \ie the distributions of noisy and clean images, requires two discriminative networks, making the design sub-optimal. 
In this paper, we adopt normalizing flows~\cite{RealNVP,Glow}, an invertible generative model, to learn the distributions and design the denoising algorithm.

The advantages of normalizing flows are reflected in three aspects. First, its invertibility ensures one-to-one mapping between images and their latent representations~\cite{liu2020deep}, 
ensuring that the manipulation on the latent representation corresponds to modifying the original input image. 
Second, it is capable of transforming complex distributions to isotropic distributions without losing information~\cite{liu2020deep}. Thus, we can obtain the accurate noisy distribution and also restore the clean images more precisely (see \autoref{fig:Visual_Front}). Last, it lets the encoder and the decoder share weights, making the model size much smaller and the training more efficient.

For the second challenge, we take advantage of the characteristics of the latent variables, which follow a distribution of $N(\bm{0}, \bm{I})$ and thus the dimensions are independent to each other. We assume that these dimensions can be disentangled into two groups, \ie some of the dimensions encode clean images while the others correspond to noise. If we set the noise dimensions to constants, such as $\bm{0}$, the joint distribution of the clean representations and new noise codes will be the same as the marginal distribution of the clean images. The denoised images can be obtained by passing the new latent representations to the reverse pass of the network.

The contributions of our work are listed below.
\begin{itemize}
    \item We rethink the image denoising task and present a distribution-learning based denoising framework.
    \item We propose a Flow Based Image Denoising Neural Network (\ourmodel{}). Unlike the widely used feature-learning based CNNs in this area, \ourmodel{} learns the distribution of noisy images instead of low-level features.
    \item We present a disentanglement method to obtain the distribution of clean ground truth from the noisy distribution, without making any assumptions on noise or employing the priors of images.
    \item We achieve competitive performance in removing synthetic noise from category-specific images and remote sensing images. For real noise, we also verify our denoising capacity by achieving a new state-of-the-art result on the real-world SIDD dataset.
\end{itemize}

\section{Related Work}
\subsection{Recent trends of image denoising}
\textbf{Traditional Methods.}
Traditional denoising methods usually construct an optimization scheme, modeling the distributions of noise or the priors of natural images as penalties or constraints.
The widely used natural image priors include sparsity~\cite{5459452}, total variation~\cite{TV,2019_TV_multi_channel}, non-local similarity~\cite{1467423,4154791} and external statistical prior~\cite{7410393,6126278}.
NLM~\cite{1467423} computes a weighted average of non-local similar patches to denoise images. The weights are calculated by the Euclidean distance between pixels.
BM3D~\cite{BM3D} employs the structure similarity of patches in a transform domain, achieving excellent accuracy on denoising additive white gaussian noise (AWGN). 

However, most traditional methods are designed to tackle generic natural images. 
Very few works study category-specific image denoising and consider the class-specific priors while designing algorithms.
CSID~\cite{CSID} is the first to adopt external similar clean patches to facilitate denoising category-specific object images. They formulate an optimization problem using the priors in the transform domain. The objective consists of a Gaussian fidelity term that incorporates the category-specific information and a low-rank term that fortifies the similarity between noisy and external similar clean patches. They achieve superior denoising accuracy in removing noise from category-specific images.
Nonetheless, a common problem that lies in most of these traditional model-driven methods is that they require noise levels as input. These methods usually implement various hard thresholds to deal with different noise levels. However, the noise level is usually unavailable, and we can only do blind denoising in practice, limiting the application of these methods.

\textbf{Deep Learning Methods.}
Deep learning denoising methods learn models from a large number of clean and noisy image pairs with CNNs, without providing image priors manually. The rapid progress of these methods has been seen in recent years, promoting the denoising effect significantly. The notable DnCNN~\cite{DnCNN} achieves good results on AWGN removal. After that, RIDNet~\cite{Anwar2019RIDNET} brings attention to denoising models, boosting the denoising performance further. VDN~\cite{VDN} makes assumptions on the distribution of clean images and noise, deriving a new form of evidence lower bound observation (ELBO) under the variational inference framework as the training objective. These CNN based denoising methods learn low-level features in the network to restore the details of clean images.

There have also been a few attempts in designing category-specific denoising networks recently, for example,~\cite{8418389} proposes a class-aware CNN based denoising method. The authors use a classifier to classify the noisy image into the supported classes first and then exploit the pre-trained class-specific denoising models for denoising. For each of the supported classes, the denoising model is
pre-trained on the images from the same classes of ImageNet~\cite{ILSVRC15}. The denoising architecture they proposed is a feature-learning based CNN.
However, for category-specific images, the feature learning based denoising methods usually enforce the pixels of denoised images to be close to the clean ones but ignore the underlying distribution of the specific category. Thus, over-smoothed regions and artifacts are seen in restored images, degrading the visual effects of denoising. 
As far as we know, we are the first to conduct image denoising with distribution learning and disentanglement.

\subsection{Flow Based Invertible Networks}
We employ normalizing flows based invertible neural networks to learn the distributions.
Normalizing flows~\cite{NICE} are models for computing complex distributions accurately. 
By applying a sequence of invertible transformations to transform a simple prior distribution into a complex distribution, the complex distribution's exact log-likelihood can be computed. 

The key design concept of normalizing flows is invertibility, ensuring the mapping between an input and its output is one-to-one. Therefore, to estimate the probability density of image $\y$, we can alternatively achieve the same purpose by measuring the probability density of the counterpart latent variable $\z \sim \ngaussian$.
Estimating the probability density of $\y$ through using the probabilities of $\z$ requires taking the variations of metric spaces into consideration. Consequently, we have: 
\begin{equation}
        p(\y) =p(\z)\big|\det(\frac{\partial f^{-1}(\z)}{\partial \z})\big|
        =p(\z)\big|\det(\frac{\partial f(\z)}{\partial \z})\big|^{-1}, 
\label{Eq:density}
\end{equation}
where $\z = f(\y)$ and $\y = f^{-1}(\z)$. $f$ is the invertible function learned by normalizing flows. 

To reduce the complexity of computing the determinants of Jacobian matrices, special designs are proposed in NICE~\cite{NICE} and Real NVP~\cite{RealNVP} to make each flow module have a triangular Jacobian matrix. Glow~\cite{Glow} extends the channel permutation methods in these two models and proposes Invertible $1\times1$ convolutional layers. These models are usually used in image generation, demonstrating superior generation quality of natural images.

So far, few studies apply invertible networks to image denoising.
Noise Flow~\cite{NoiseFlow} employs Glow~\cite{Glow} to learn the distribution of real-world noise and generate real noisy images for data augmentation. Extra information, such as raw images, ISO, and camera-specific parameters, is required during noisy image generation. 
Different from these studies, we are the first to exploit normalizing flows to learn the distribution of noisy images and disentangle the clean representations to restore images. 
\section{Our Method}
In this section, we explain the design concept of \ourmodel. Then we introduce the detailed components of the network architecture. The objective function, as well as some training details, are also presented. 

\begin{figure}
\centering
\includegraphics[width=0.5\textwidth]{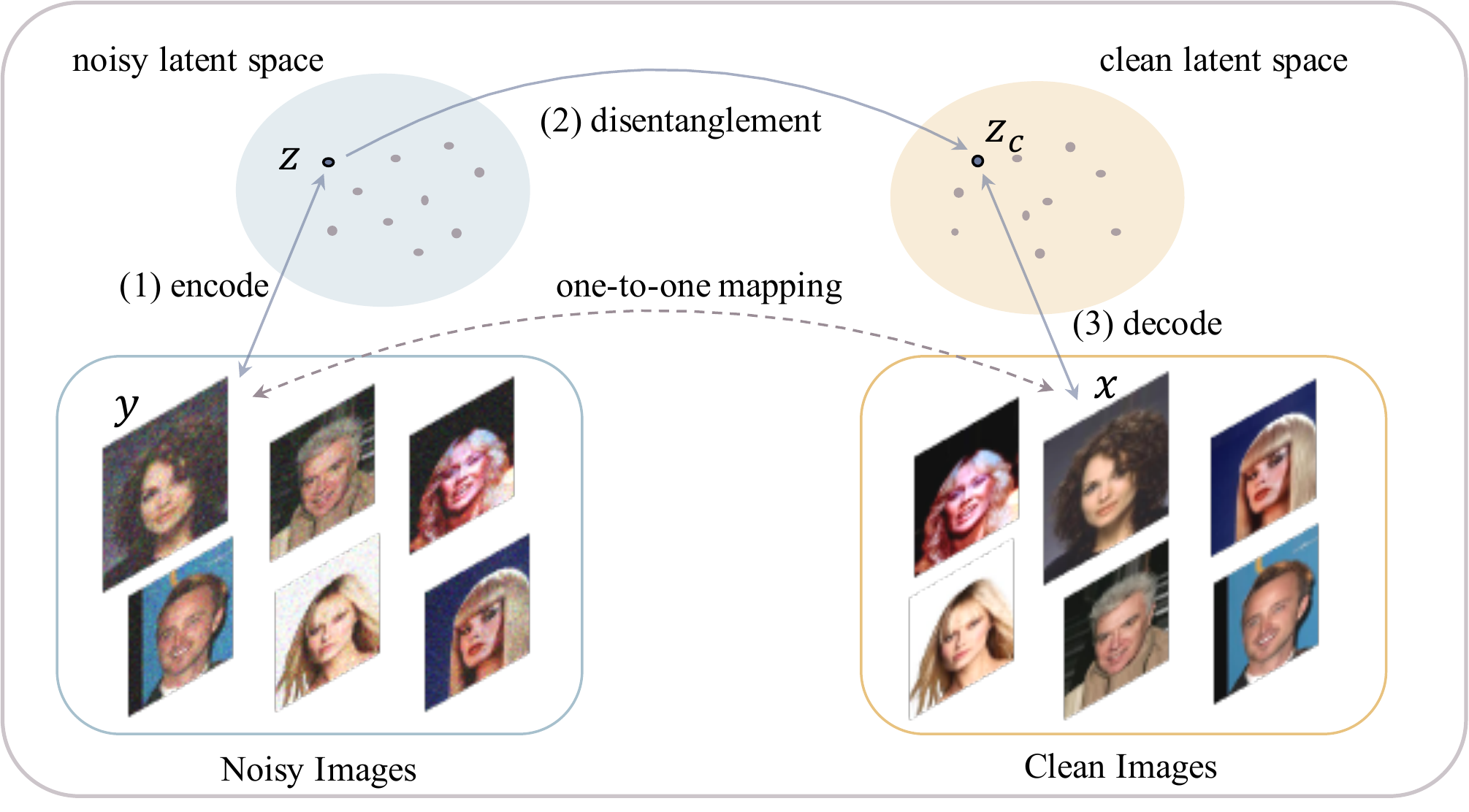}
\caption{The framework of distribution learning and disentanglement based image denoising.}   
\label{fig:framework}
\end{figure}

\begin{figure*}
\centering
\includegraphics[width=\textwidth]{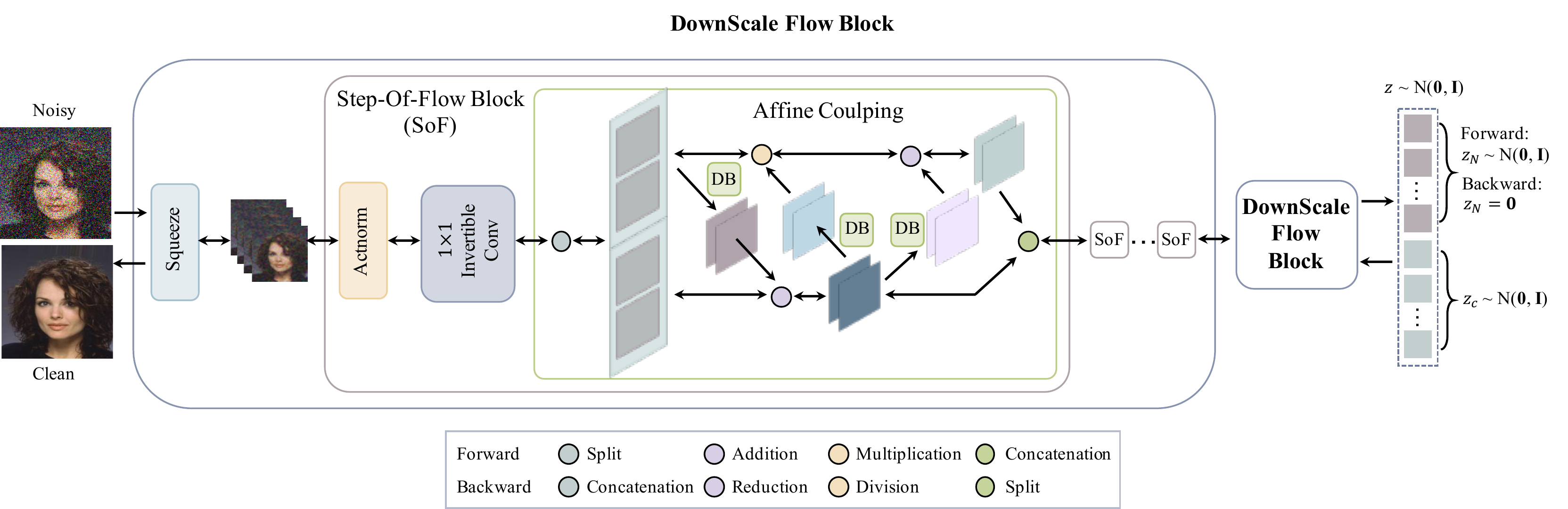}
\caption{Our \ourmodel{} Network Architecture. \ourmodel{} consists of several invertible DownScale Flow Blocks. The forward pass encodes the corrupted image to latent variables $\bm{z} = (\bm{z_N}, \bm{z_C}) \sim \ngaussian$. The latent representations of noise $\bm{z_N}$ are set to $\bm{0}$, combined with the clean latent variables $\bm{z_C}$ as a new latent representation $\hatz$. The backward pass decodes $\hatz$ to the denoised image.}
\label{fig:NetArchi}
\end{figure*}

\subsection{Concept of Design}
We rethink the image denoising task from the perspective of distribution learning and disentanglement. 
Suppose the noisy image is $\y$ and the corresponding clean ground truth is $\x$. The noise $\n = \y-\x $. 
We have: $p(\y) = p(\x, \n) =  p(\x) p(\n | \x)$. That is, the distribution of the noisy images $p(\y)$ is a joint distribution of clean images and noise. 
The clean representation can be achieved if we can disentangle the clean and noise representations from $p(\y)$. Then, the clean images can be restored with the disentangled clean representation.

A framework of this scheme is presented in~\autoref{fig:framework}, which contains three steps: i) learn the distribution of noisy images by encoding $\y$ to a noisy latent representation $\z$, ii) disentangle the clean representation $\zc$ from $\z$, and iii) restore the clean image by decoding $\zc$ to the clean image space. To ensure the denoising effect, the mappings between $\y$ and $\z$, $\zc$ and $\x$ should be one-to-one.

An invertible normalizing flow based network is employed to learn the distribution of noisy images $p(\y)$, transforming $\y$ to latent variables $\z$ following an simple prior distribution $\ngaussian$. Thus,
\begin{equation}
    p(\y) = p(\x) p(\n | \x) =p(\z)\big|\det(\frac{\partial f(\z)}{\partial \z})|\big|^{-1}, 
\end{equation}
where $f(\cdot)$ is the model learned by the network. The dimensions of $\z$ are independent of each other.

We assume $\z$ can be disentangled and some of the dimensions of $\z$ encode the distribution of clean images (denoted as $\zc$) and the remaining embeds noise (denoted as $\zn$). The clean image $\x$ can be restored through the following transformation:
\begin{equation}
    p(\x) =p(\zc)\big|\det(\frac{\partial f(\z)}{\partial \zc})\big|^{-1}. 
\label{Equ:p_x}
\end{equation}

However, how to obtain $\zc$ with $\z$ is not so obvious.
We propose a way of disentanglement by setting $\zn=\bm{0}$, that is,  
\begin{equation}
    \hatz = {\bf m} \odot \z,
\end{equation}
where ${\bf m}$ is a mask which is 1 in the dimensions for clean variables and 0 in those for noise. $\hatz$ is a new latent code which only contains the clean representations. $\odot$ denotes the element-wise product.
Thus, we have $p({\zn=\bm{0}})=1$ and the distribution of $\hatz$ becomes
\begin{equation}
    p(\hatz) =p(\zc)p(\zn)=p(\zc).
\end{equation}

Then the clean image can be obtained via Eq.~\eqref{Equ:p_x}.

\subsection{Network Architecture}
The details of our \ourmodel{} architecture are presented in this section. \ourmodel{} is composed of several invertible DownScale Flow Blocks, as shown in~\autoref{fig:NetArchi}. Each block consists of a Squeeze layer to downscale the latent representations followed by several Step-Of-Flow Blocks to perform distribution transformation. The details of each layer are described below.

\begin{figure}
\centering
\includegraphics[width=0.48\textwidth]{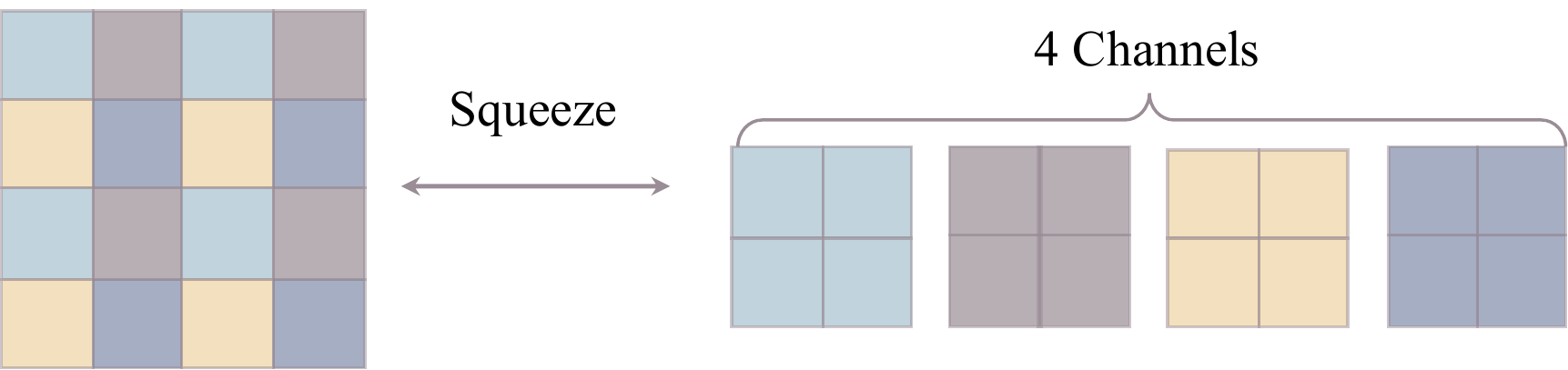}
\caption{The Squeeze operation downscales latent representations according to a checkerboard pattern.}   
\label{fig:Transform}
\vspace{-5mm}
\end{figure}

\textbf{Squeeze.} The Squeeze layers take every other element of the intermediate latent variables, resulting in new downscaled latent representations with quadruple channels, as illustrated in \autoref{fig:Transform}.

\textbf{Actnorm.} The Actnorm layers apply the affine transformation on latent variables, as illustrated in Eq.~\eqref{Equ: Actnorm}. 
\begin{equation}\label{Equ: Actnorm}
    \hii = \sa \odot \hi + \biasa,
\end{equation}
where $\hi$ and $\hii$ are the intermediate latent representations during transformation. $\sa$ and $\biasa$ are the scale and translation parameters separately. $\odot$ is the Hadamard product of tensors. The reverse operation of the Actnorm layer is
\begin{equation}\label{Equ: Actnorm}
    \hi = (\hii - \biasa) / \sa,
\end{equation}
$\sa$ and $\biasa$ are initialized to make each channel of the representations have zero mean and unit variance, like the normalization operation. However, during training, this operation is different from the widely used normalization methods. Specifically, $\sa$ and $\biasa$ are updated through back-propagation, without any further constraints on the mean and variance of the latent variables. Employing the Actnorm layers is able to improve the training stability and performance. 

\textbf{Invertible $1\times1$ Convolutional Layers.} Different from ordinary convolutional layers, 
we use the invertible $1\times1$ convolutional layers, which are designed for normalizing flows to support invertibility.
The operation in these layers can be represented as
\begin{equation}\label{Equ: Invertible}
    \hii = \textbf{W} \hi, 
\end{equation}
where $\textbf{W}$ is a square matrix which is initialized randomly.
Its reverse function is
\begin{equation}\label{Equ: Invertible}
    \hi = \textbf{W}^{-1} \hii. 
\end{equation}
These layers are used to permute different channels of latent representations.

\textbf{Affine Coupling Layers.} The Affine Coupling layers capture the correlations among spatial dimensions~\cite{NICE,RealNVP}. The forward operations include:
\begin{gather*}\label{Equ: Affine Coupling_1}
    \hi^{a}, \hi^{b} = \text{Split}(\hi),  \\
    \hii^{a} = \hi^{a}+ \text{g}_1(\hi^b), \\ 
    \hii^{b} = \text{g}_2(\hii^a) \odot \hi^{b}+ \text{g}_3(\hii^a), \\ 
    \hii = \text{Concat}(\hii^{a}, \hii^{b}), 
\end{gather*}
where $\text{Split}(\cdot)$ and $\text{Concat}(\cdot)$ operate along channel dimensions. $\text{Split}(\cdot)$ splits $\hi$ into two tensors $\hi^{a}$ and $\hi^{b}$. $\text{Concat}(\cdot)$ concatenates two tensors $\hii^{a}$, $\hii^{b}$ channel-wise to obtain $\hii$. $\text{g}_i(\cdot)$ ($i=1, 2, 3$) is neural network. The reverse operations are:
\begin{gather*}\label{Equ: Affine Coupling_1}
    \hii^{a}, \hii^{b} = \text{Split}(\hii),  \\
    \hi^{b} = (\hii^{b} - \text{g}_3(\hii^a))/\text{g}_2(\hii^a) , \\ 
    \hi^{a} = \hii^{a} - \text{g}_1(\hi^b), \\ 
    \hi = \text{Concat}(\hi^{a}, \hi^{b}). 
\end{gather*}

The operations in the second and third row turn $+$ into $-$ and $\odot$ into $/$. $\text{g}_i(\cdot)$ ($i$=1, 2, 3) can be any neural network. Following~\cite{ardizzone2019guided} and~\cite{xiao2020invertible}, we employ Dense Block (DB) in our network as $\text{g}_i(\cdot)$.

\subsection{Objective Function}
Our objective function consists of two components, the distribution learning loss to encode the input noisy image $\y$ into latent code $\z$, and the reconstruction loss to restore the corresponding clean image $\x$ with clean code $\zc$. The details of these two losses are as below.

\textbf{Distribution Learning Loss.}
\begin{equation}\label{loss:enc}
    L_{\text{dis}} = - \log p(\y) = - \big(\log p_{z}(\z)+\sum_{i=1}^{L} \log (\det|\frac{\partial f_{i}}{\partial \hi}|)\big)\;,
\end{equation}
where $L$ is the number of invertible layers in \ourmodel{} and $f_{i}$ is the function learned by each layer. $p_{z}(\z) \sim \ngaussian$ and $\z = (\zc, \zn)$.
To reconstruct the clean image, we set $\zn = \bm{0}$ and achieve a new latent representation $\hatz = (\zc, \bm{0})$, which lies in a subspace of $\z$.

\textbf{Reconstruction Loss.}
$\hatz$ is passed through the reverse network to restore the clean ground truth.
\begin{equation}\label{loss:rec}
    L_{\text{rec}} = ||f^{-1}(\hatz) - \x||_1\;.
\end{equation}

\textbf{Total Loss.} The total objective function we use during training is:
\begin{equation}
    L = \lambda_{1} L_{\text{dis}} + \lambda_{2} L_{\text{rec}}\;,
\end{equation}
where $\lambda_{1}$ and $\lambda_{2}$ are the weights for the two loss components.

\begin{figure}[t]
\centering
\begin{subfigure}{.23\textwidth}
  \centering
  \includegraphics[width=\textwidth]{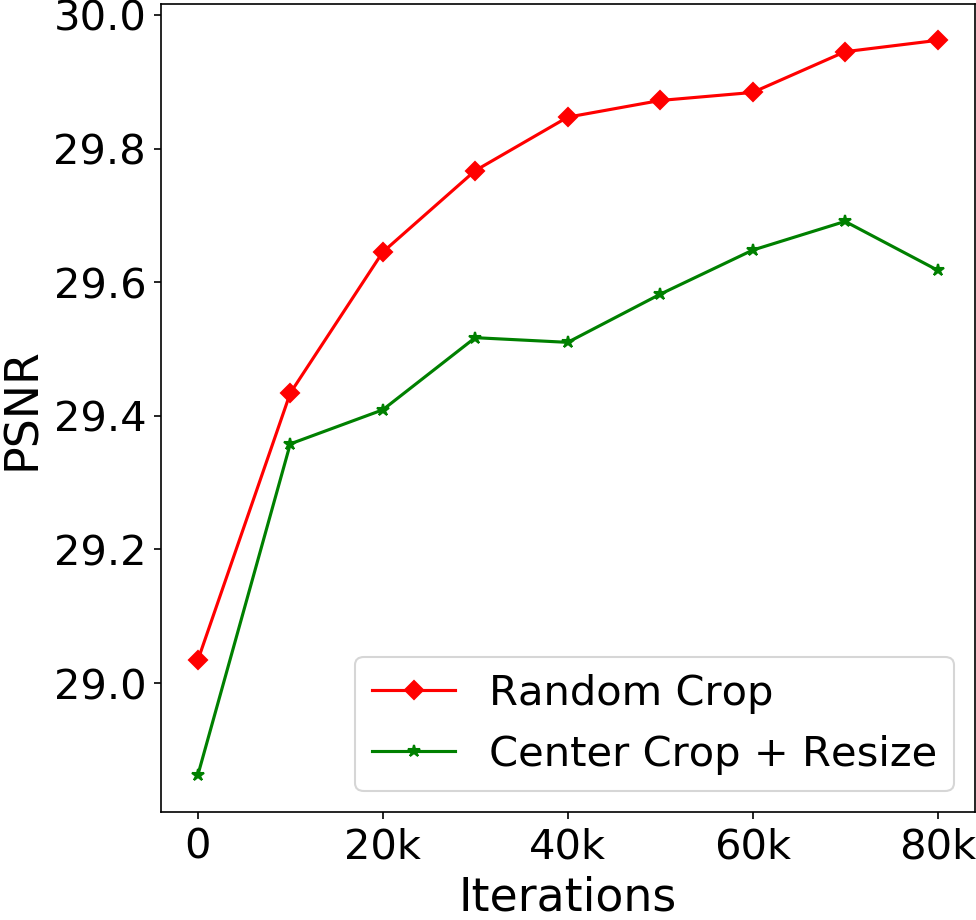} 
  \caption{Crop Method}
  \label{subfig:crop}
\end{subfigure}
\begin{subfigure}{.23\textwidth}
  \centering
  \includegraphics[width=\textwidth]{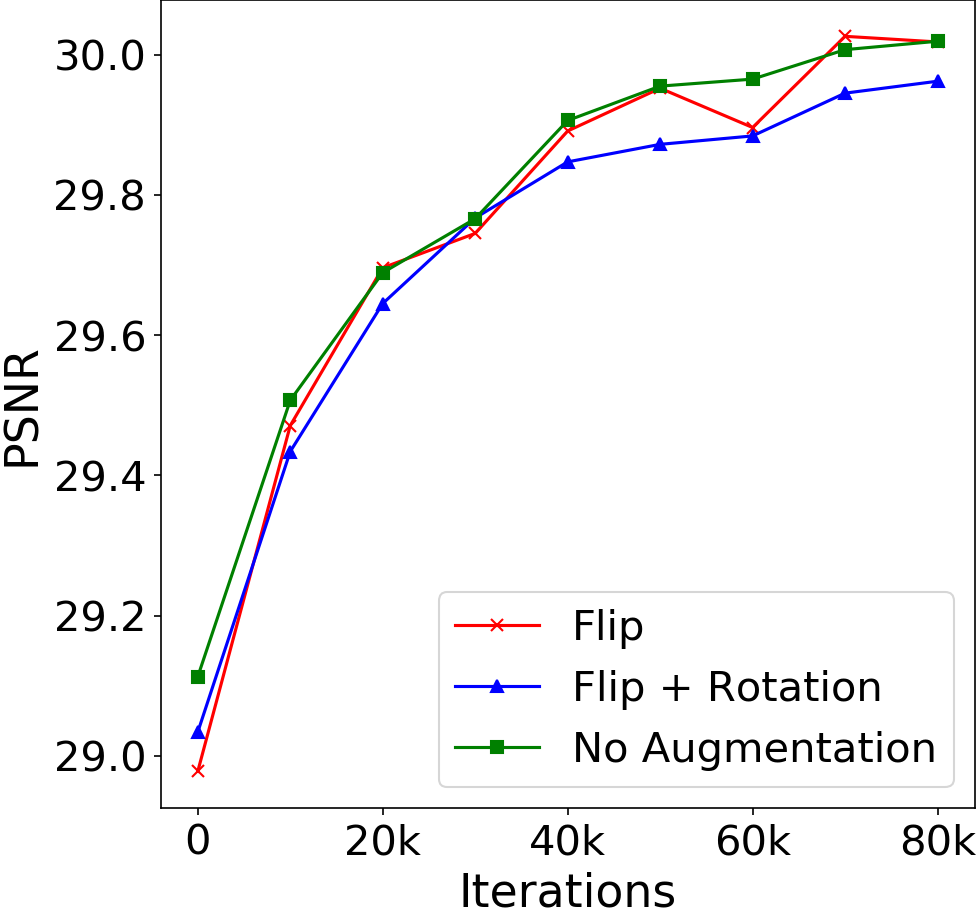} 
  \caption{Data Augmentation}
  \label{subfig:augmentation}
\end{subfigure}
\caption{Training with different processing strategies. (a) Illustrates the difference between random and center crop accompanied with resizing. (b) The validation curves using different data augmentation methods. All the models are trained on CelebA~\cite{CelebA} with $\sigma=50$.}
\end{figure}

\subsection{Data Preprocessing}
Since \ourmodel{} is the first distribution learning-based denoising network, we explore different data preprocessing techniques to demonstrate how to make the best use of it.

\textbf{Random vs. Center Crop $\&$ resize.}
The widely used training strategy in feature learning based CNN denoising networks is to randomly crop patches from the training dataset to learn features. However, it is not obvious whether the random crop strategy is also superior in distribution learning based networks. Take face image denoising as an example; if we center crop the face region and resize it into an appropriate size, we will get a downscaled face image, following a similar distribution as the face image test set. Intuitively, the center crop with the resizing method will facilitate the network to learn the distribution better and achieve superior denoising results. 

Thus, we compare training with random crop and center crop with resizing, illustrating the curves of the validation results on~\autoref{subfig:crop}. Contrary to our intuition, cropping training patches randomly outperforms the center crop with resizing consistently and significantly.  
Therefore, we apply random crops during the training \ourmodel{}.

\textbf{Data Augmentation vs. No Data Augmentation.}
Feature learning based networks usually employ horizontal and vertical flip and rotation with 90, 180, 270 degrees for data augmentation. However, these methods will bring in unrealistic patches, compromising the learning of distributions. For example, if we rotate a patch of a face image, we may get patches with the eyes under the mouth or on the mouth's left side, which is impossible in real face images. 
Although data augmentation can lead to better generalizability for discriminative models, it may bring noise when learning the distributions. 

We train three models for comparison: one with flip and rotation as data augmentation, one with only flip as data augmentation, and the other is trained without any augmentation. The validation results are shown in ~\autoref{subfig:augmentation}. The results verify our concern that inappropriate data augmentation such as rotation introduces noise to the distribution model, resulting in a lower denoising accuracy shown in the blue curve. Training with only flip as data augmentation achieves almost the same results as without data augmentation, but the latter is more stable during training.
The potential reason might be that the horizontal flip also generates realistic images for face images, while the vertical flip creates impossible face images, making the training unstable.
Thus, to avoid unrealistic samples, we train our distribution learning based networks without data augmentation when the training set is large enough to learn the distribution. If the training set is small, we only conduct data augmentation that will not generate unrealistic patches.

\section{Experiment}
We perform thorough experiments to demonstrate the effectiveness of our method.
We first apply \ourmodel{} to denoise category-specific images. Since category-specific images usually have similar patterns in all the images, such as similar facial contours and features in human faces, their distribution is easier to learn than random nature images.  
The experiment is then extended to denoising more difficult remote sensing images, which contain diverse terrain patterns, such as mountains or forests, following intricate distributions. Finally, we investigate our capacity to remove noise, which follows complicated distribution in the real noise dataset.
Further details are provided about the datasets, training strategies, and qualitative and quantitative results.

\begin{figure*}[t]
\centering
\begin{subfigure}{.13\textwidth}
  \centering
  \includegraphics[width=.99\linewidth]{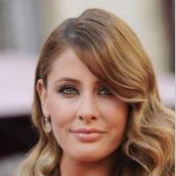}  
\end{subfigure}
\begin{subfigure}{.13\textwidth}
  \centering
  \includegraphics[width=.99\linewidth]{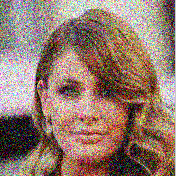}  
\end{subfigure}
\begin{subfigure}{.13\textwidth}
  \centering
  \includegraphics[width=.99\linewidth]{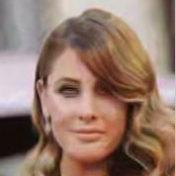}  
\end{subfigure}
\begin{subfigure}{.13\textwidth}
  \centering
  \includegraphics[width=.99\linewidth]{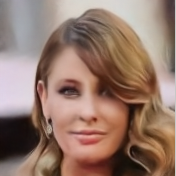}  
\end{subfigure}
\begin{subfigure}{.13\textwidth}
  \centering
  \includegraphics[width=.99\linewidth]{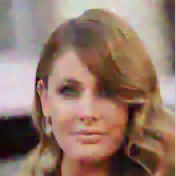}
\end{subfigure}
\begin{subfigure}{.13\textwidth}
  \centering
  \includegraphics[width=.99\linewidth]{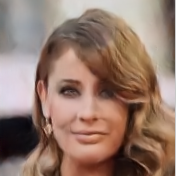}
\end{subfigure}
\begin{subfigure}{.13\textwidth}
  \centering
  \includegraphics[width=.99\linewidth]{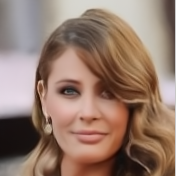}
\end{subfigure}
\vspace{1mm}\\

\begin{subfigure}{.13\textwidth}
  \centering
  \includegraphics[width=.99\linewidth]{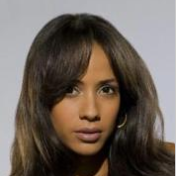}  
\end{subfigure}
\begin{subfigure}{.13\textwidth}
  \centering
  \includegraphics[width=.99\linewidth]{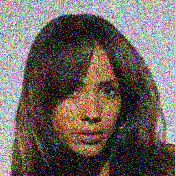}  
\end{subfigure}
\begin{subfigure}{.13\textwidth}
  \centering
  \includegraphics[width=.99\linewidth]{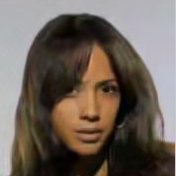}  
\end{subfigure}
\begin{subfigure}{.13\textwidth}
  \centering
  \includegraphics[width=.99\linewidth]{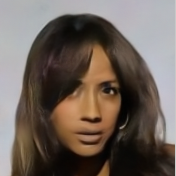}  
\end{subfigure}
\begin{subfigure}{.13\textwidth}
  \centering
  \includegraphics[width=.99\linewidth]{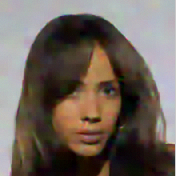}
\end{subfigure}
\begin{subfigure}{.13\textwidth}
  \centering
  \includegraphics[width=.99\linewidth]{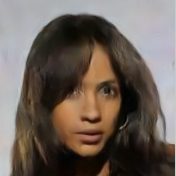}
\end{subfigure}
\begin{subfigure}{.13\textwidth}
  \centering
  \includegraphics[width=.99\linewidth]{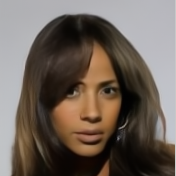}
\end{subfigure}
\vspace{1mm}\\

\begin{subfigure}{.13\textwidth}
  \centering
  \includegraphics[width=.99\linewidth]{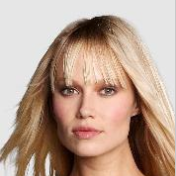}  
\end{subfigure}
\begin{subfigure}{.13\textwidth}
  \centering
  \includegraphics[width=.99\linewidth]{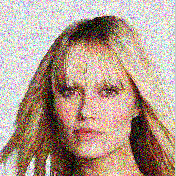}  
\end{subfigure}
\begin{subfigure}{.13\textwidth}
  \centering
  \includegraphics[width=.99\linewidth]{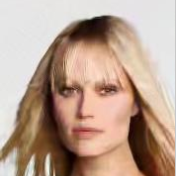}  
\end{subfigure}
\begin{subfigure}{.13\textwidth}
  \centering
  \includegraphics[width=.99\linewidth]{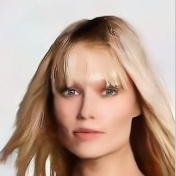}  
\end{subfigure}
\begin{subfigure}{.13\textwidth}
  \centering
  \includegraphics[width=.99\linewidth]{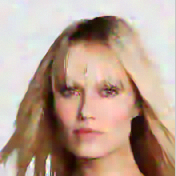}
\end{subfigure}
\begin{subfigure}{.13\textwidth}
  \centering
  \includegraphics[width=.99\linewidth]{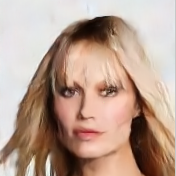}
\end{subfigure}
\begin{subfigure}{.13\textwidth}
  \centering
  \includegraphics[width=.99\linewidth]{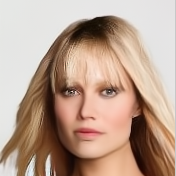}
\end{subfigure}
\vspace{1mm}\\

\begin{subfigure}{.13\textwidth}
  \centering
  \includegraphics[width=.99\linewidth]{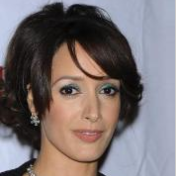}  
\end{subfigure}
\begin{subfigure}{.13\textwidth}
  \centering
  \includegraphics[width=.99\linewidth]{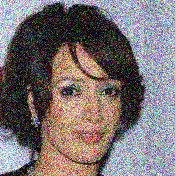}  
\end{subfigure}
\begin{subfigure}{.13\textwidth}
  \centering
  \includegraphics[width=.99\linewidth]{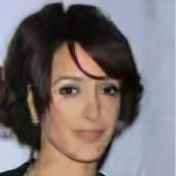}  
\end{subfigure}
\begin{subfigure}{.13\textwidth}
  \centering
  \includegraphics[width=.99\linewidth]{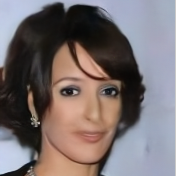}  
\end{subfigure}
\begin{subfigure}{.13\textwidth}
  \centering
  \includegraphics[width=.99\linewidth]{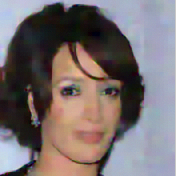}
\end{subfigure}
\begin{subfigure}{.13\textwidth}
  \centering
  \includegraphics[width=.99\linewidth]{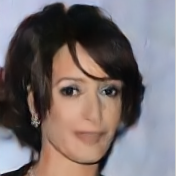}
\end{subfigure}
\begin{subfigure}{.13\textwidth}
  \centering
  \includegraphics[width=.99\linewidth]{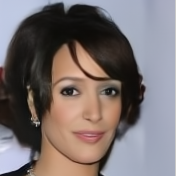}
\end{subfigure}
\vspace{1mm}\\

\begin{subfigure}{.13\textwidth}
  \centering
  \caption*{GT} 
\end{subfigure}
\begin{subfigure}{.13\textwidth}
  \centering
  \caption*{Noisy} 
\end{subfigure}
\begin{subfigure}{.13\textwidth}
  \centering
  \caption*{BM3D} 
\end{subfigure}
\begin{subfigure}{.13\textwidth}
  \centering
  \caption*{DnCNN}   
\end{subfigure}
\begin{subfigure}{.13\textwidth}
  \centering
  \caption*{EPLL} 
\end{subfigure}
\begin{subfigure}{.13\textwidth}
  \centering
  \caption*{IRCNN} 
\end{subfigure}
\begin{subfigure}{.13\textwidth}
  \centering
  \caption*{\ourmodel{} (Ours)} 
\end{subfigure}
\caption{The image denoising results of \ourmodel{} on CelebA dataset with $\sigma=50$ against competitive methods. Our network produces results close to the ground-truth without any kind of deformation and artifact. The effects are best viewed with zooming in.}
\label{fig:Visual_CelebA}
\end{figure*}

\subsection{Experimental Settings}

\subsubsection{Training Details}
\ourmodel{} with two DownScale Flow blocks and eight SoF blocks in each Flow block is exploited in our experiment, where ADAM~\cite{kingma2014adam} is applied as an optimizer. The learning rate is initialized as $2 \times 10^{-4}$ and halved after every 50K iterations. To evaluate the methods, we employ Peak Signal-to-Noise Ratio (PSNR) as the evaluation metric.

\subsubsection{Datasets}
Next, we provide information about the category specific, remote sensing and real world datasets.

\vspace{1mm}
\noindent
\textbf{Category-Specific Datasets:} We investigate the capacity of \ourmodel{} in removing AWGN on three category-specific datasets: faces, flowers, and birds. 
\begin{itemize}
\item  CelebA~\cite{CelebA} is a large human face dataset containing 202,599 face images. We use the 162770 training images for training and 19867 validation images for testing. The training images are cropped into $64 \times 64$ patches randomly as the network's input at the training stage. Since the training set is large enough, we do not apply any data augmentation during training. 
    
\item  Flower Dataset~\cite{Flower} contains 102 categories of flowers, including 1020 training images, 1020 validation images, and 6149 test images. To better learn the distribution of flowers, we change the dataset's partition and use the 6149 images as the training set and the remaining 2040 images as the test set. The training images are randomly cropped into patches with a size of $128 \times 128$. Flipping and rotation are employed as data augmentation. 

\item  CUB-200 Dataset~\cite{Bird} includes 11,788 bird images, covering 200 categories of birds. We use 5989 images as the training set and 5790 images as the test set. The training images are cropped into $128 \times 128$ patches with random flipping as data augmentation during training.
\end{itemize}

\vspace{1mm}
\noindent
\textbf{Remote Sensing Datasets:} 
We attempt to denoise two remote sensing datasets (RICE1 and RICE2~\cite{RICE}) with AWGN added to explore our capability when the distribution of ground truth images becomes complex. The datasets contain 500 and 450 pairs of images respectively, each with a size of $512\times512$. We randomly crop patches of size $64\times64$ from the images for training and add AWGN with $\sigma=30, 50$, and $70$ to get noise-free and noisy pairs, respectively. Random flipping, as well as rotation, are utilized for data augmentation.

\vspace{1mm}
\noindent
\textbf{Real Noisy Datasets:}
Finally, we verify~\ourmodel's effectiveness in removing real noise, which follows a complex distribution. Real image noise can result from photon shot noise, fixed pattern noise, dark current, readout noise, quantization noise, \etc during the imaging process~\cite{Real_Noise_Source}. We conduct real noise removal on the dataset SIDD~\cite{SIDD_2018_CVPR}, which is taken by five smartphone cameras with small apertures and sensor sizes. The medium SIDD dataset contains 320 clean and noisy pairs. Patches with a size of $144 \times 144$ are randomly cropped for training. Flipping and rotation are adopted for augmentation.

\begin{figure*}[t]
\centering
\begin{subfigure}{.13\textwidth}
  \centering
  \includegraphics[width=.99\linewidth]{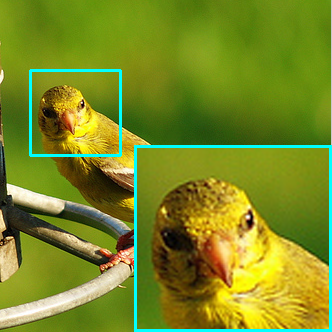}  
\end{subfigure}
\begin{subfigure}{.13\textwidth}
  \centering
  \includegraphics[width=.99\linewidth]{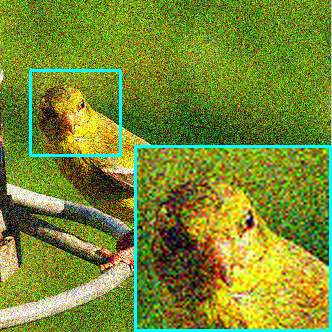}  
\end{subfigure}
\begin{subfigure}{.13\textwidth}
  \centering
  \includegraphics[width=.99\linewidth]{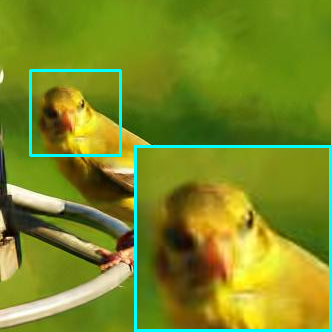}  
\end{subfigure}
\begin{subfigure}{.13\textwidth}
  \centering
  \includegraphics[width=.99\linewidth]{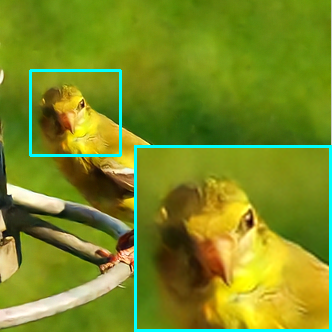}  
\end{subfigure}
\begin{subfigure}{.13\textwidth}
  \centering
  \includegraphics[width=.99\linewidth]{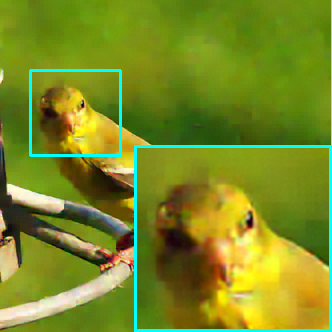}
\end{subfigure}
\begin{subfigure}{.13\textwidth}
  \centering
  \includegraphics[width=.99\linewidth]{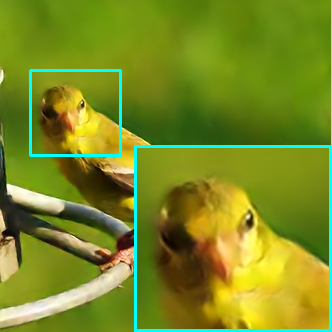}
\end{subfigure}
\begin{subfigure}{.13\textwidth}
  \centering
  \includegraphics[width=.99\linewidth]{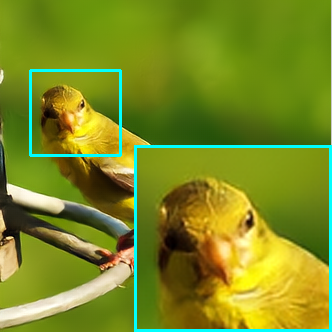}
\end{subfigure}
\vspace{1mm}\\

\begin{subfigure}{.13\textwidth}
  \centering
  \includegraphics[width=.99\linewidth]{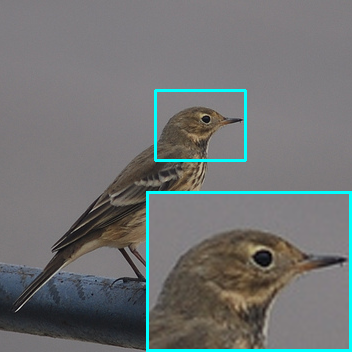}  
\end{subfigure}
\begin{subfigure}{.13\textwidth}
  \centering
  \includegraphics[width=.99\linewidth]{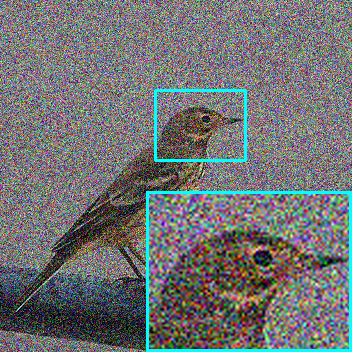}  
\end{subfigure}
\begin{subfigure}{.13\textwidth}
  \centering
  \includegraphics[width=.99\linewidth]{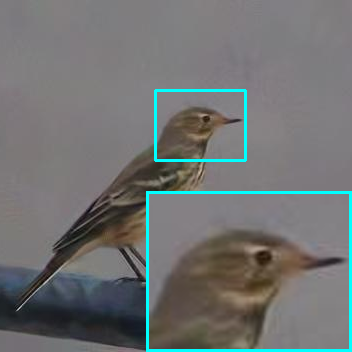}  
\end{subfigure}
\begin{subfigure}{.13\textwidth}
  \centering
  \includegraphics[width=.99\linewidth]{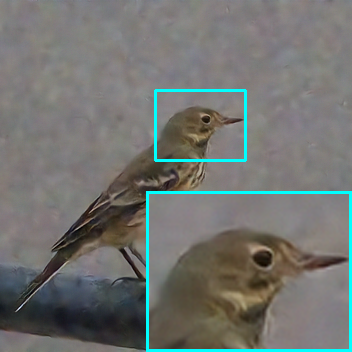}  
\end{subfigure}
\begin{subfigure}{.13\textwidth}
  \centering
  \includegraphics[width=.99\linewidth]{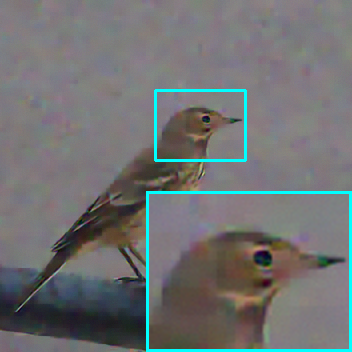}
\end{subfigure}
\begin{subfigure}{.13\textwidth}
  \centering
  \includegraphics[width=.99\linewidth]{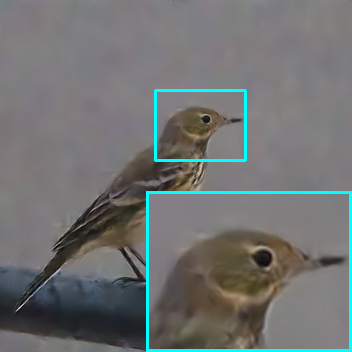}
\end{subfigure}
\begin{subfigure}{.13\textwidth}
  \centering
  \includegraphics[width=.99\linewidth]{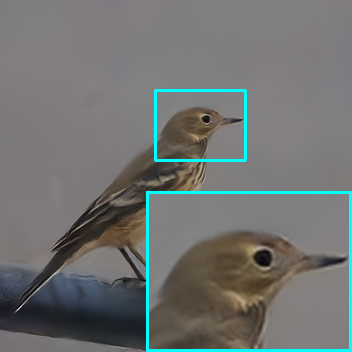}
\end{subfigure}
\vspace{1mm}\\

\begin{subfigure}{.13\textwidth}
  \centering
  \includegraphics[width=.99\linewidth]{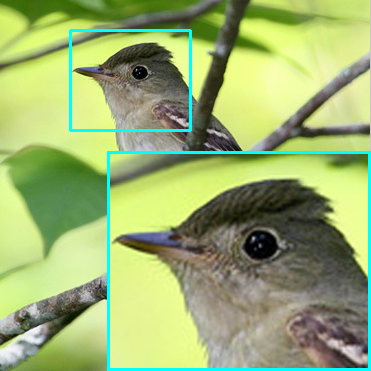}  
\end{subfigure}
\begin{subfigure}{.13\textwidth}
  \centering
  \includegraphics[width=.99\linewidth]{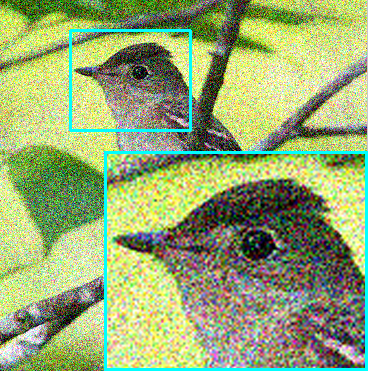}  
\end{subfigure}
\begin{subfigure}{.13\textwidth}
  \centering
  \includegraphics[width=.99\linewidth]{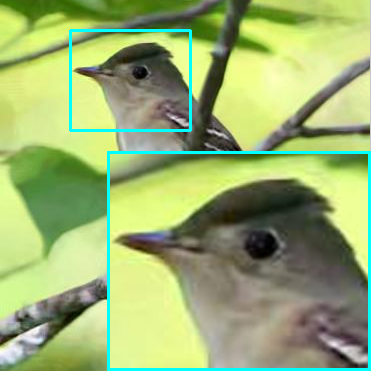}  
\end{subfigure}
\begin{subfigure}{.13\textwidth}
  \centering
  \includegraphics[width=.99\linewidth]{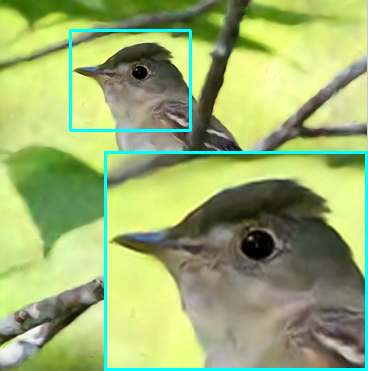}  
\end{subfigure}
\begin{subfigure}{.13\textwidth}
  \centering
  \includegraphics[width=.99\linewidth]{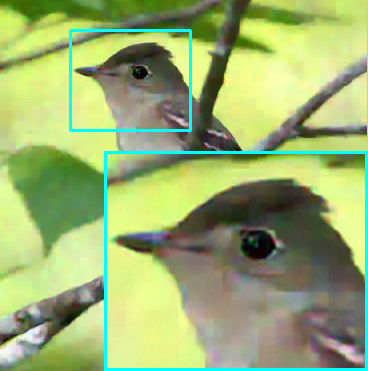}
\end{subfigure}
\begin{subfigure}{.13\textwidth}
  \centering
  \includegraphics[width=.99\linewidth]{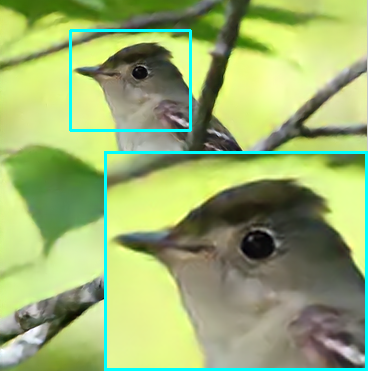}
\end{subfigure}
\begin{subfigure}{.13\textwidth}
  \centering
  \includegraphics[width=.99\linewidth]{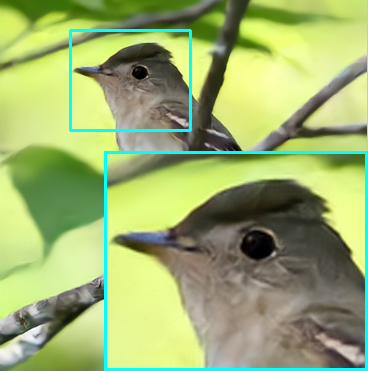}
\end{subfigure}
\vspace{1mm}\\

\begin{subfigure}{.13\textwidth}
  \centering
  \includegraphics[width=.99\linewidth]{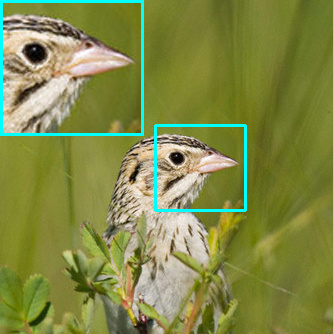}  
\end{subfigure}
\begin{subfigure}{.13\textwidth}
  \centering
  \includegraphics[width=.99\linewidth]{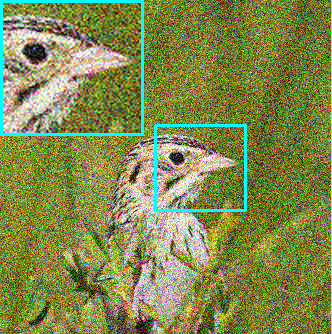}  
\end{subfigure}
\begin{subfigure}{.13\textwidth}
  \centering
  \includegraphics[width=.99\linewidth]{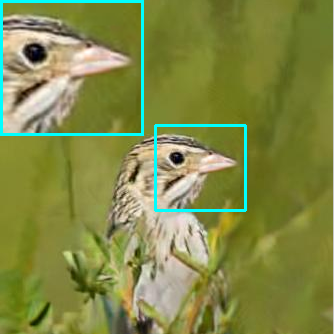}  
\end{subfigure}
\begin{subfigure}{.13\textwidth}
  \centering
  \includegraphics[width=.99\linewidth]{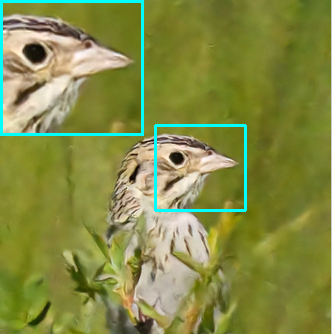}  
\end{subfigure}
\begin{subfigure}{.13\textwidth}
  \centering
  \includegraphics[width=.99\linewidth]{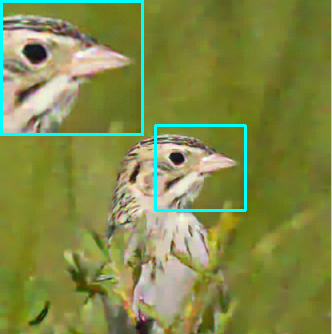}
\end{subfigure}
\begin{subfigure}{.13\textwidth}
  \centering
  \includegraphics[width=.99\linewidth]{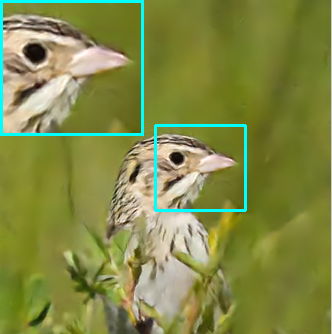}
\end{subfigure}
\begin{subfigure}{.13\textwidth}
  \centering
  \includegraphics[width=.99\linewidth]{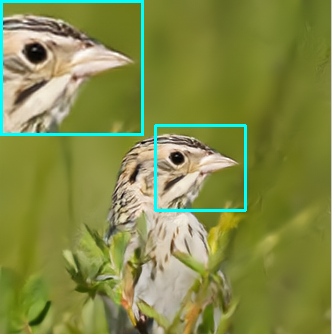}
\end{subfigure}
\vspace{1mm}\\

\begin{subfigure}{.13\textwidth}
  \centering
  \caption*{GT} 
\end{subfigure}
\begin{subfigure}{.13\textwidth}
  \centering
  \caption*{Noisy} 
\end{subfigure}
\begin{subfigure}{.13\textwidth}
  \centering
  \caption*{BM3D} 
\end{subfigure}
\begin{subfigure}{.13\textwidth}
  \centering
  \caption*{DnCNN}   
\end{subfigure}
\begin{subfigure}{.13\textwidth}
  \centering
  \caption*{EPLL} 
\end{subfigure}
\begin{subfigure}{.13\textwidth}
  \centering
  \caption*{IRCNN} 
\end{subfigure}
\begin{subfigure}{.13\textwidth}
  \centering
  \caption*{\ourmodel{} (Ours)} 
\end{subfigure}
\caption{The comparison of denoising results on CUB-200 dataset having $\sigma=50$. Our method removes artifacts and noise, providing clean edges and textures.}
\label{fig:Visual_Bird}
\end{figure*}

\subsection{Ablation Study}
\textbf{Number of Flow and SoF Blocks.}
We study the denoising effects of employing different numbers of Flow and SoF blocks in~\ourmodel. We train models on CelebA~\cite{CelebA} with $\sigma=50$ AWGN added. The results of the 50Kth iteration on the validation set are reported in~\autoref{Tab:blocks}.
In general, given the same number of Flow blocks, the more SoF blocks in each of the Flow blocks, the higher the denoising accuracy. However, the improvement is not significant when we have 3 Flow blocks.
On the other hand, with the same number of SoF blocks in each Flow block, increasing Flow block numbers from 1 to 2 improves the performance. Nevertheless, further increment results in similar accuracy when SoF=4 and even slightly decreases when SoF=8. Thus, we adopt 2 Flow blocks with 8 SoF blocks contained in our experiment.

\begin{table}[]
\centering
\caption{Comparisons on the denoising accuracy of different numbers of DownScale Blocks and Invertible Blocks.}
\begin{tabular}{c|c|c|c}
\hline\hline
\multicolumn{2}{l|}{\multirow{2}{*}{PSNR}} & \multicolumn{2}{c}{SoF Blocks}                 \\ \cline{3-4} 
\multicolumn{2}{l|}{}& \multicolumn{1}{c|}{num = 4} & \multicolumn{1}{c}{num = 8} \\ \hline
\multirow{3}{*}{Flow Blocks} & num = 1 & 29.59 & 29.86 \\ \cline{2-4} 
& num = 2 & 29.87 & \textbf{30.00} \\ \cline{2-4} 
& num = 3 & 29.89 & 29.90 \\ \hline
\hline
\end{tabular}
\label{Tab:blocks}
\end{table}

\begin{table}
\centering
\caption{Comparisons on different proportions of the dimensions of $\zc$ in $\z$.} 
\begin{tabular}{cccccc}
\hline
\hline
dim($\zc$) & 1/8 & 1/4 & 1/2 & 3/4 & 7/8 \\ 
\hline
 PSNR & 29.81 & 30.00 & 30.00 & \textbf{30.19} & 30.18 \\ \hline
\hline
\end{tabular}
\label{Tab:dim}
\end{table}

\begin{table*}[]
\centering
\caption{The quantitative comparison of removing synthetic noise on three category-specific datasets. Our \ourmodel{} outperforms the other competitive methods on all of the three datasets for various noise levels.} 
\begin{tabular}{l|c|c|c|c|c|c|c|c}
\hline\hline
Dataset & $\sigma$ & BM3D~\cite{BM3D} & EPLL~\cite{EPLL} & IRCNN~\cite{IRCNN} & REDNet~\cite{REDNet} & DnCNN~\cite{DnCNN} & FFDNet~\cite{zhang2018ffdnet} & \ourmodel{} (Ours) \\ \hline
& 15 & 35.46 & 33.29 & 35.20 & 35.23 & 35.04 & 35.14 & \textbf{35.74} \\
& 25 & 32.80 & 30.81 & 32.62 & 32.68 & 32.63 & 32.40 & \textbf{32.95} \\
& 50 & 29.46 & 27.65 & 29.24 & 29.56 & 29.57 & 29.44 & \textbf{30.29} \\ 
\multirow{-4}{*}{CelebA~\cite{CelebA}} & Blind & -- & -- & 31.92 & 33.16 & 32.17 & 32.20 & \textbf{33.52} \\ \hline
& 15 & 37.20 & 35.41 & 36.83 & 36.93 & 36.73 & 36.49 & \textbf{37.38} \\
& 25 & 34.73 & 32.92 & 34.47 & 34.75 & 34.17 & 33.89 & \textbf{34.82} \\
& 50 & 31.38 & 29.58 & 30.8 & 31.34 & 30.38 & 30.74 & \textbf{31.71} \\
\multirow{-4}{*}{Flower~\cite{Flower}} & Blind & -- & -- & 34.91 & 34.57 & 34.55 & 34.61 & \textbf{35.18}\\ \hline
& 15 & 35.08 & 33.31 & 35.14 & 35.16 & 35.21 & 34.86 & \textbf{35.30} \\
& 25 & 32.59 & 30.83 & 32.71 & 32.80 & 32.45 & 32.33 & \textbf{32.94} \\
& 50 & 29.32 & 27.61 & 29.28 & 29.72 & 28.87 & 28.61 & \textbf{29.79} \\
\multirow{-4}{*}{CUB-200~\cite{Bird}} & Blind & -- & -- & 32.49 & 33.18 & 32.89 & 33.07 & \textbf{33.20} \\\hline
\hline
\end{tabular}
\label{Tab:Class_Specific AWGN Denoise}
\end{table*}

\textbf{The split of $\zc$ and $\zn$.}
We also study the effects of different dimension numbers of $\zc$ (\ie dim($\zc$)). Models are trained with dim($\zc$) = 1/8, 1/4, 1/2, 3/4, and 7/8 dim($\z$) separately, and the validation results of the 50Kth iteration are reported in~\autoref{Tab:dim}. In general, the denoising accuracy improves with the increase of the proportion of dim($\zc$). On the other hand, the results are almost the same when dim($\zc$) = 3/4 and 7/8 dim($\z$), illustrating that extending the dimensions of $\zc$ will not boost the denoising performance further. Thus, we use dim($\zc$) = 3/4 dim($\z$) in our experiment.

\subsection{Category-Specific Image Denoising}

\textbf{Quantitative Results.} 
In Table~\ref{Tab:Class_Specific AWGN Denoise}, we report numeric values for the three category-specific datasets  with added AWGN with the levels of $\sigma=15, 25$ and $50$. Compared with other competitive methods in synthetic noise removal, \ourmodel{} achieves the highest PSNR on all the datasets and noise levels.

\begin{figure*}[ht]
\centering
\begin{subfigure}{.13\textwidth}
  \centering
  \includegraphics[width=.99\linewidth]{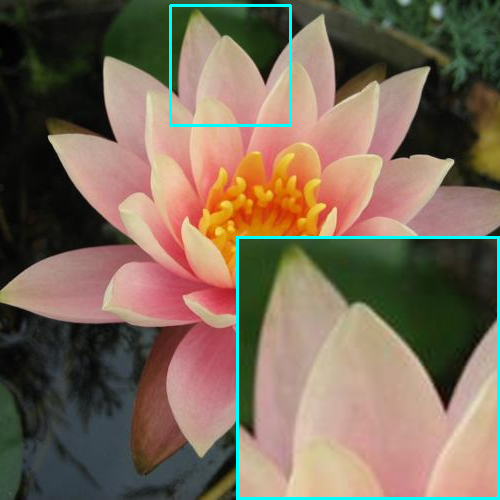}  
\end{subfigure}
\begin{subfigure}{.13\textwidth}
  \centering
  \includegraphics[width=.99\linewidth]{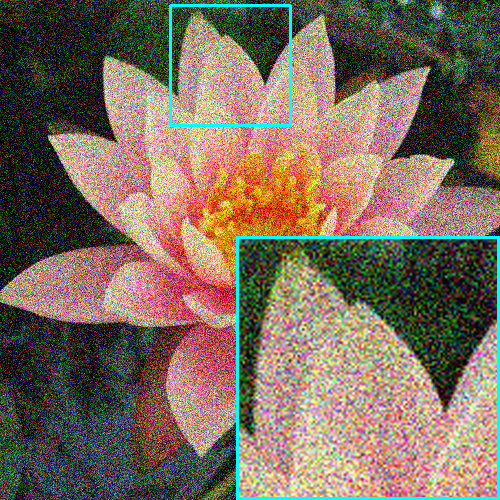}  
\end{subfigure}
\begin{subfigure}{.13\textwidth}
  \centering
  \includegraphics[width=.99\linewidth]{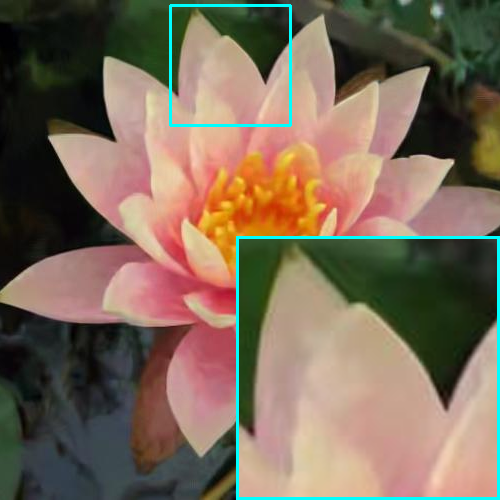}  
\end{subfigure}
\begin{subfigure}{.13\textwidth}
  \centering
  \includegraphics[width=.99\linewidth]{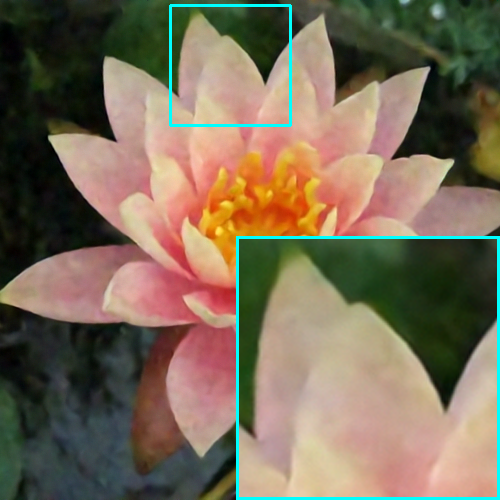}  
\end{subfigure}
\begin{subfigure}{.13\textwidth}
  \centering
  \includegraphics[width=.99\linewidth]{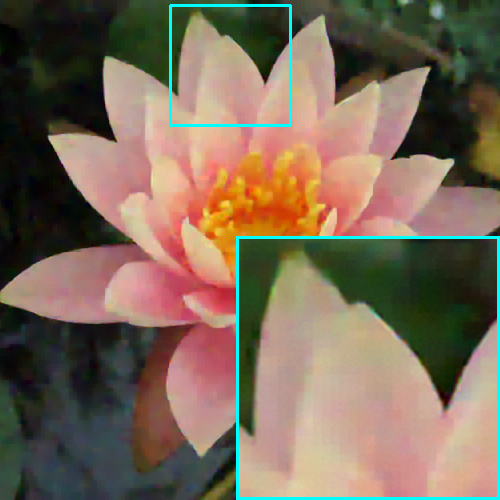}
\end{subfigure}
\begin{subfigure}{.13\textwidth}
  \centering
  \includegraphics[width=.99\linewidth]{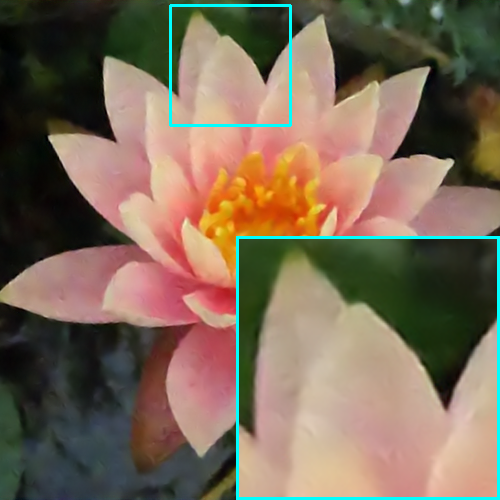}
\end{subfigure}
\begin{subfigure}{.13\textwidth}
  \centering
  \includegraphics[width=.99\linewidth]{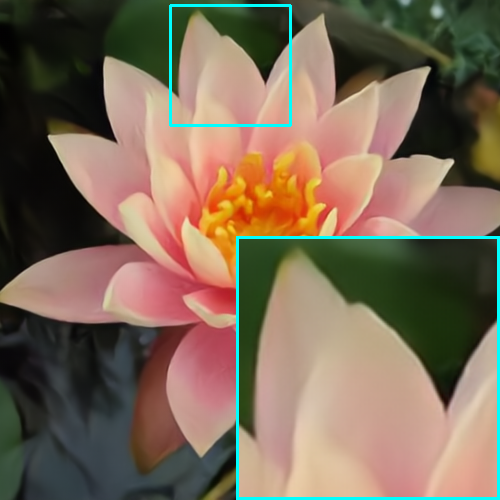}
\end{subfigure}
\vspace{1mm}\\

\begin{subfigure}{.13\textwidth}
  \centering
  \includegraphics[width=.99\linewidth]{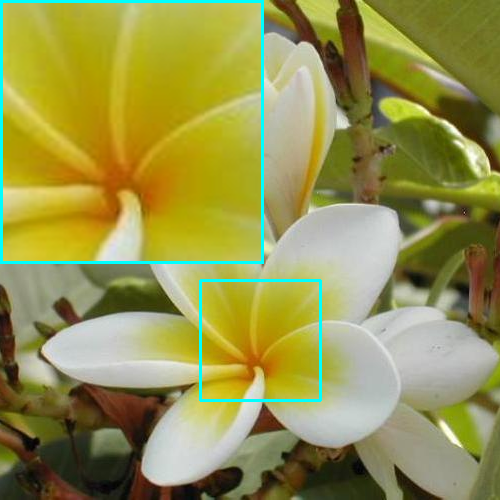}  
\end{subfigure}
\begin{subfigure}{.13\textwidth}
  \centering
  \includegraphics[width=.99\linewidth]{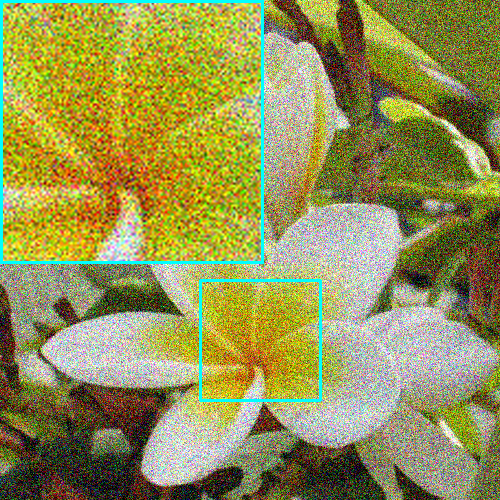}  
\end{subfigure}
\begin{subfigure}{.13\textwidth}
  \centering
  \includegraphics[width=.99\linewidth]{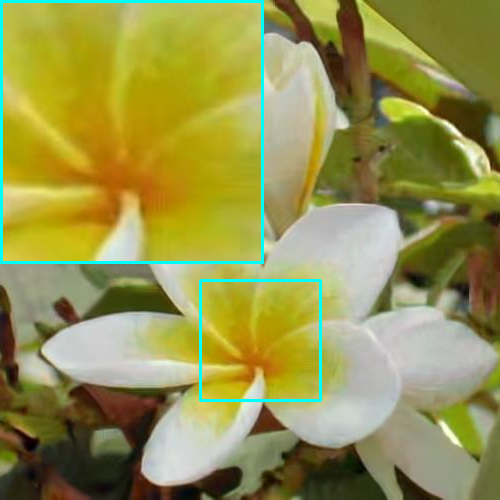}  
\end{subfigure}
\begin{subfigure}{.13\textwidth}
  \centering
  \includegraphics[width=.99\linewidth]{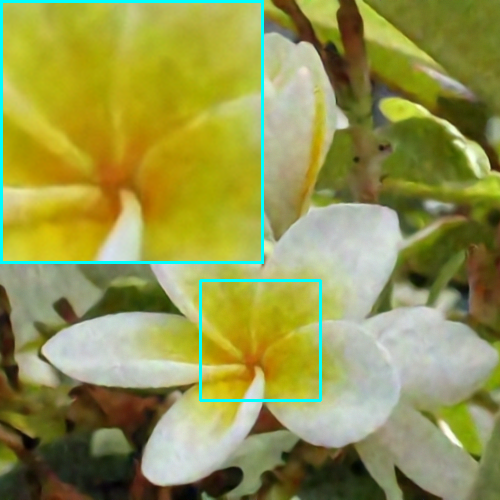}  
\end{subfigure}
\begin{subfigure}{.13\textwidth}
  \centering
  \includegraphics[width=.99\linewidth]{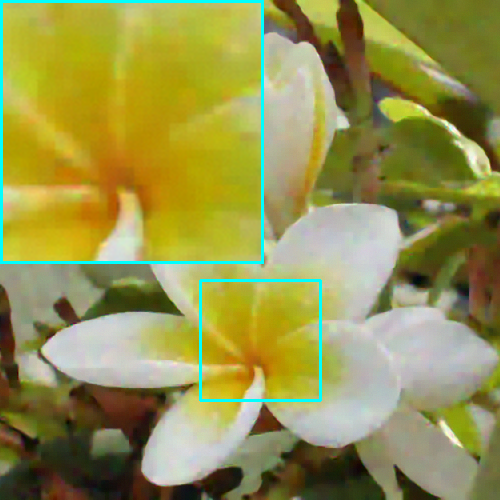}
\end{subfigure}
\begin{subfigure}{.13\textwidth}
  \centering
  \includegraphics[width=.99\linewidth]{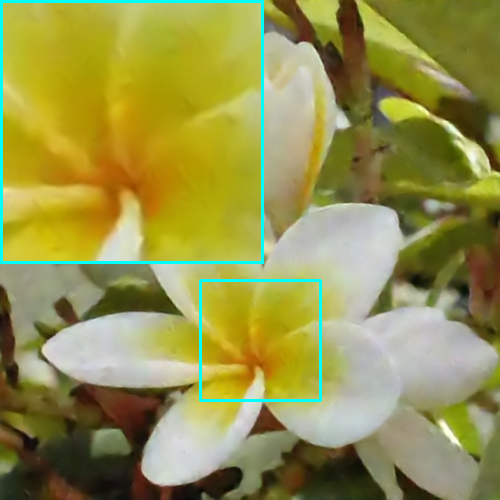}
\end{subfigure}
\begin{subfigure}{.13\textwidth}
  \centering
  \includegraphics[width=.99\linewidth]{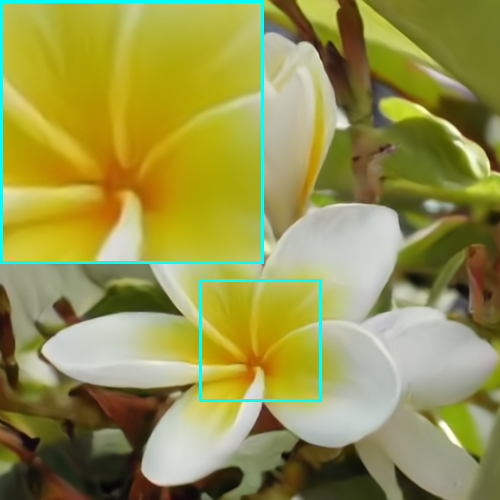}
\end{subfigure}
\vspace{1mm}\\

\begin{subfigure}{.13\textwidth}
  \centering
  \includegraphics[width=.99\linewidth]{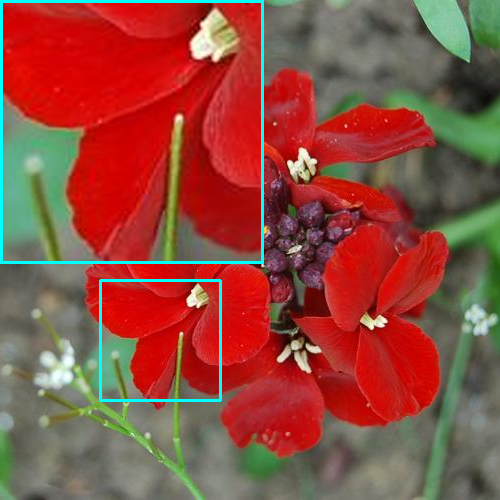}  
\end{subfigure}
\begin{subfigure}{.13\textwidth}
  \centering
  \includegraphics[width=.99\linewidth]{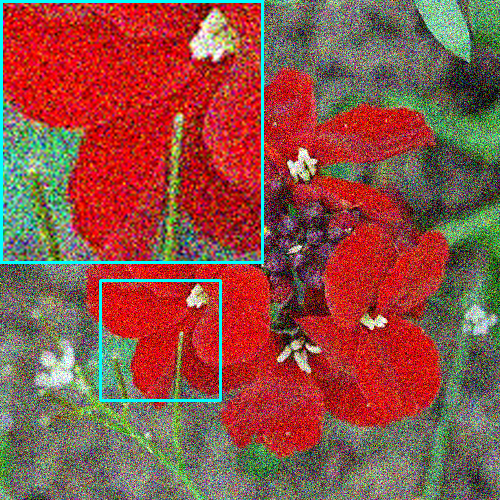}  
\end{subfigure}
\begin{subfigure}{.13\textwidth}
  \centering
  \includegraphics[width=.99\linewidth]{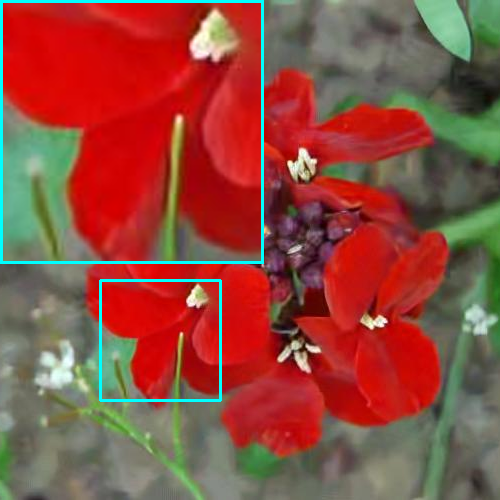}  
\end{subfigure}
\begin{subfigure}{.13\textwidth}
  \centering
  \includegraphics[width=.99\linewidth]{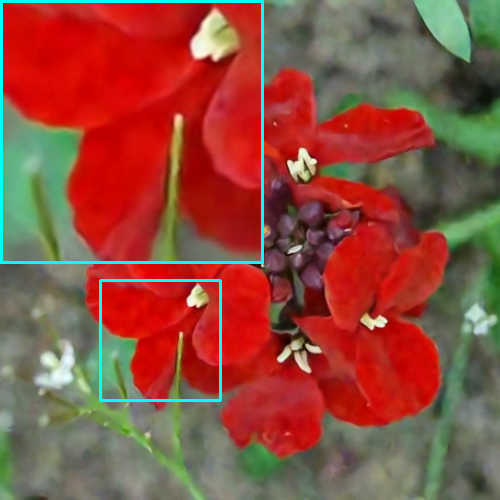}  
\end{subfigure}
\begin{subfigure}{.13\textwidth}
  \centering
  \includegraphics[width=.99\linewidth]{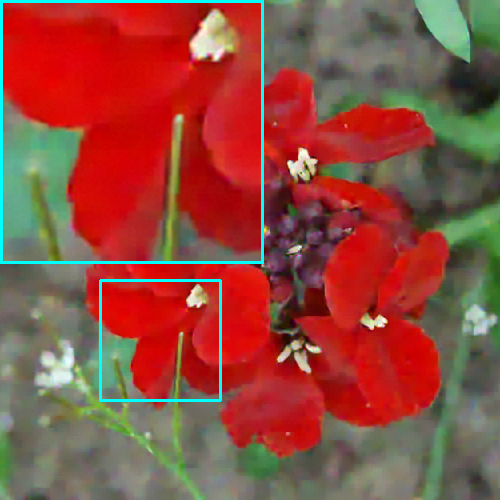}
\end{subfigure}
\begin{subfigure}{.13\textwidth}
  \centering
  \includegraphics[width=.99\linewidth]{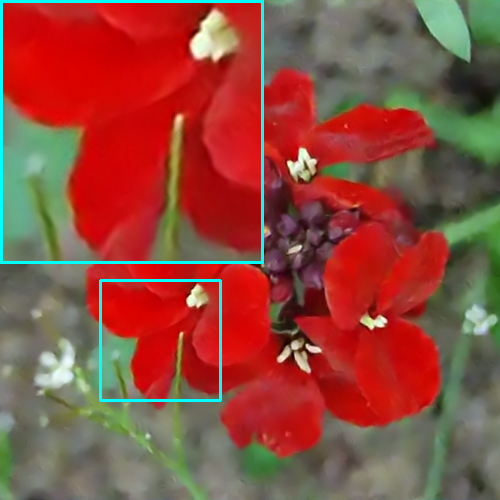}
\end{subfigure}
\begin{subfigure}{.13\textwidth}
  \centering
  \includegraphics[width=.99\linewidth]{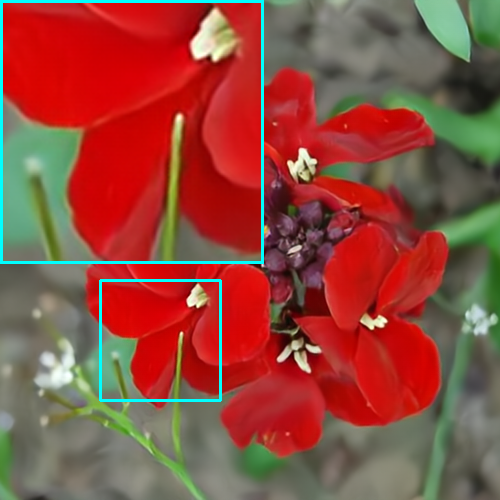}
\end{subfigure}
\vspace{1mm}\\

\begin{subfigure}{.13\textwidth}
  \centering
  \includegraphics[width=.99\linewidth]{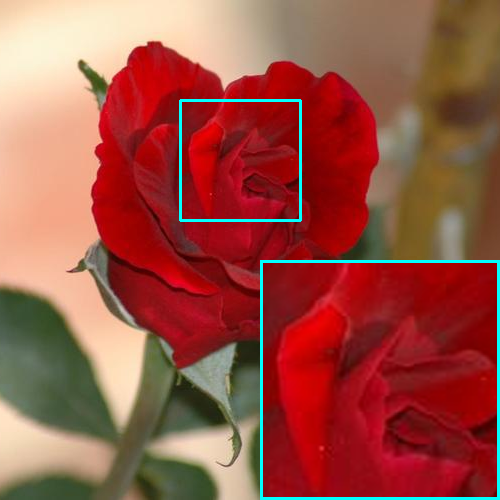}  
\end{subfigure}
\begin{subfigure}{.13\textwidth}
  \centering
  \includegraphics[width=.99\linewidth]{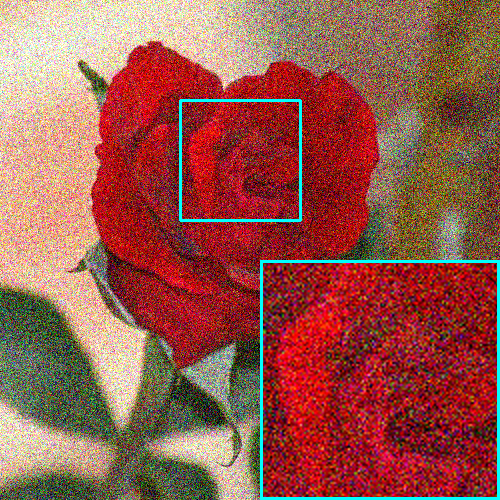}  
\end{subfigure}
\begin{subfigure}{.13\textwidth}
  \centering
  \includegraphics[width=.99\linewidth]{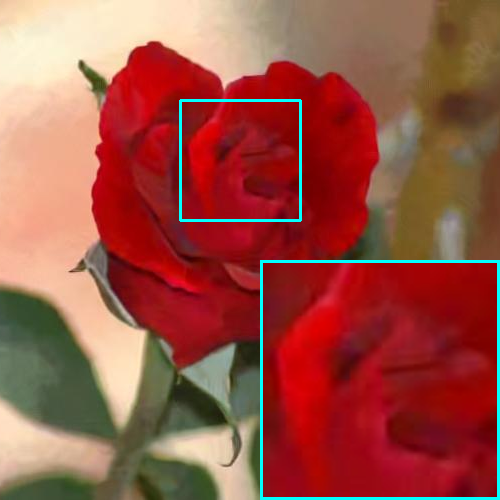}  
\end{subfigure}
\begin{subfigure}{.13\textwidth}
  \centering
  \includegraphics[width=.99\linewidth]{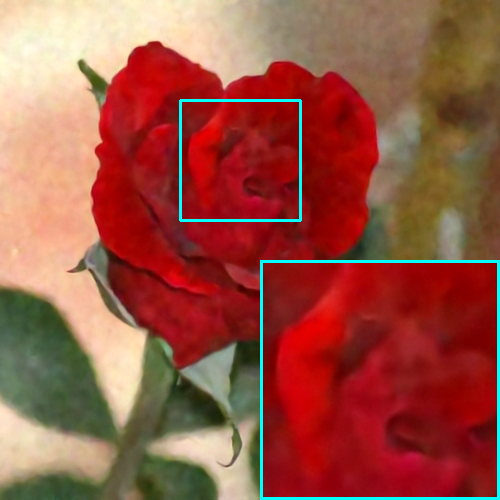}  
\end{subfigure}
\begin{subfigure}{.13\textwidth}
  \centering
  \includegraphics[width=.99\linewidth]{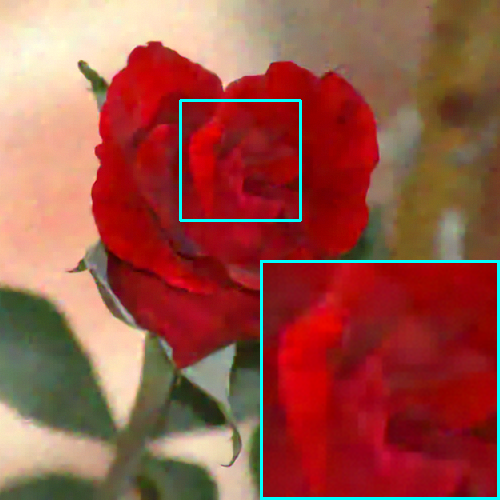}
\end{subfigure}
\begin{subfigure}{.13\textwidth}
  \centering
  \includegraphics[width=.99\linewidth]{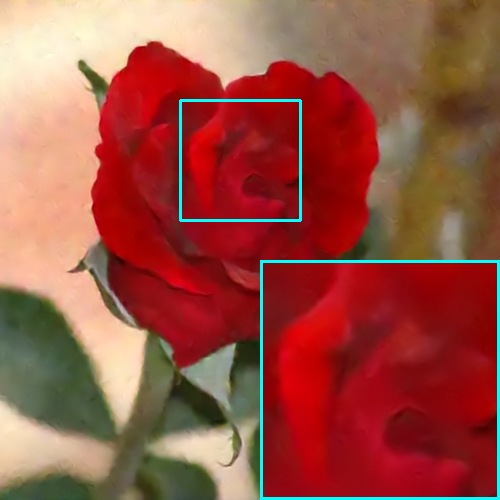}
\end{subfigure}
\begin{subfigure}{.13\textwidth}
  \centering
  \includegraphics[width=.99\linewidth]{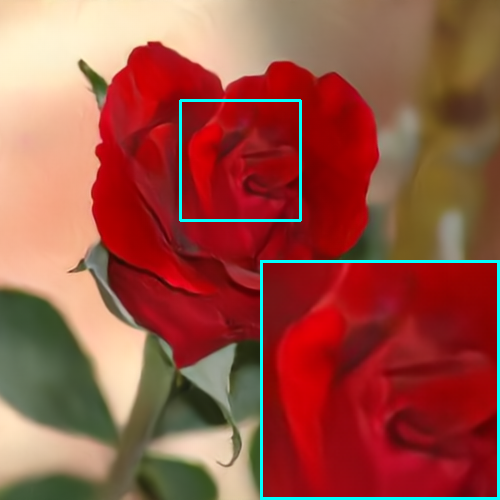}
\end{subfigure}
\vspace{1mm}\\

\begin{subfigure}{.13\textwidth}
  \centering
  \caption*{GT} 
\end{subfigure}
\begin{subfigure}{.13\textwidth}
  \centering
  \caption*{Noisy} 
\end{subfigure}
\begin{subfigure}{.13\textwidth}
  \centering
  \caption*{BM3D} 
\end{subfigure}
\begin{subfigure}{.13\textwidth}
  \centering
  \caption*{DnCNN}   
\end{subfigure}
\begin{subfigure}{.13\textwidth}
  \centering
  \caption*{EPLL} 
\end{subfigure}
\begin{subfigure}{.13\textwidth}
  \centering
  \caption*{IRCNN} 
\end{subfigure}
\begin{subfigure}{.13\textwidth}
  \centering
  \caption*{\ourmodel{} (Ours)} 
\end{subfigure}

\caption{The visual comparison on the Flower dataset for $\sigma=50$ against state-of-the-art methods.}
\label{fig:Visual_Flower}
\end{figure*}

We also employ the same \ourmodel{} for blind denoising with noise levels between [0, 55], as shown in~\autoref{Tab:Class_Specific AWGN Denoise}. The distribution of blind noise is a Gaussian Mixture Model, which is much more complicated than the Gaussian distribution with a certain noise level. 
Although traditional methods such as BM3D~\cite{BM3D} and EPLL~\cite{EPLL} are good at removing Gaussian noise, their capacity in blind denoising is unavailable due to the requirement of the noise level as input.
Comparing with other feature learning-based CNN methods, \ourmodel{} outperforms others to a large extent, exhibiting our superiority in category-specific image denoising.

\textbf{Qualitative Results.} The visual results are shown in~\autoref{fig:Visual_CelebA}, \autoref{fig:Visual_Bird} and \autoref{fig:Visual_Flower}. For CelebA, we observe that, although the other competitive methods can restore the facial contour, they lose many detailed facial features. Thus, the denoised images of these methods are blurred with artifacts. In contrast, our denoising results are much clearer and closer to the ground truth images. For the Flower and CUB-200 datasets, the foreground and background are more diverse and complicated than CelebA. Our results are clean with sharp edges (in close-up versions), while other methods have artifacts near and at the edges. This illustrates that \ourmodel{} can handle category-specific image denoising very well.

\begin{table*}[]
\centering
\caption{The performance comparison on the two remote sensing datasets.} 
\begin{tabular}{l|c|c|c|c|c|c|c|c}
\hline\hline
Dataset & $\sigma$ & EPLL~\cite{EPLL} & MemNet~\cite{Tai-MemNet-2017} & IRCNN~\cite{IRCNN} & REDNet~\cite{REDNet} & DnCNN~\cite{DnCNN} & FFDNet~\cite{zhang2018ffdnet} & \ourmodel{} (Ours) \\ \hline
& 30 & 31.95 & 31.82 & 31.12 & 29.98 & 30.69 & 22.68 & \textbf{33.08} \\
& 50 & 29.65 & 27.71 & 27.50 & 28.82 & 26.99 & 24.17 & \textbf{31.14} \\
\multirow{-3}{*}{RICE1~\cite{RICE}} & 70 & 28.29 & 27.12 & 26.53 & 26.56 & 25.04 & 23.51 & \textbf{30.03} \\\hline
& 30 & 36.05 & \textbf{36.49} & 35.83 & 33.12 & 34.68 & 34.02 & 35.93 \\
& 50 & 33.22 & 33.62 & 33.74 & 30.40 & 29.57 & 30.26 & \textbf{34.71} \\
\multirow{-3}{*}{RICE2~\cite{RICE}} & 70 & 31.63 & 31.73 & 32.43 & 27.55 & 30.81 & 28.51 & \textbf{33.98} \\ \hline
\hline
\end{tabular}
\label{Tab:RICE}
\end{table*}

\begin{figure*}[ht]
\centering
\begin{subfigure}{.16\textwidth}
  \centering
  \includegraphics[width=.99\linewidth]{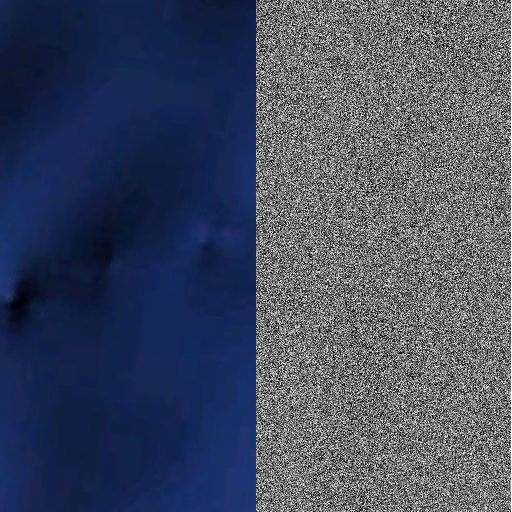} 
  \caption{Noisy}
  \label{fig:sub_noisy_RICE}
\end{subfigure}
\begin{subfigure}{.16\textwidth}
  \centering
  \includegraphics[width=.99\linewidth]{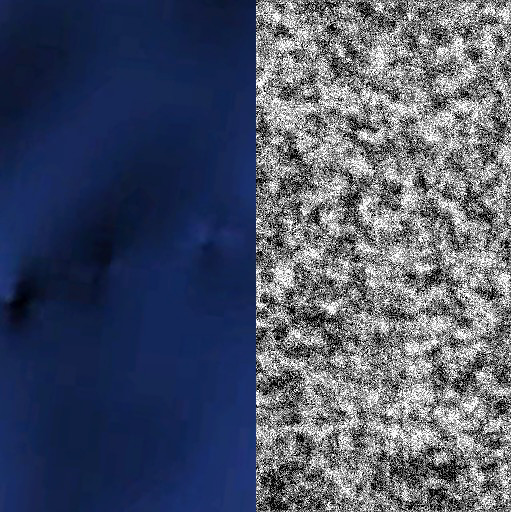}  
  \caption{DnCNN}
  \label{fig:sub_DnCNN_RICE1}
\end{subfigure}
\begin{subfigure}{.16\textwidth}
  \centering
  \includegraphics[width=.99\linewidth]{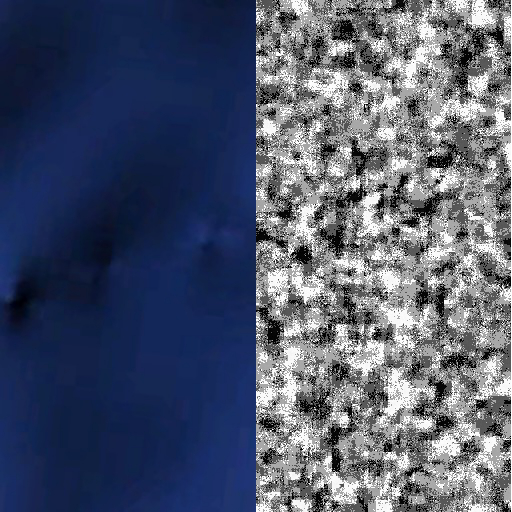}  
  \caption{EPLL}
  \label{fig:sub_EPLL_RICE}
\end{subfigure}
\begin{subfigure}{.16\textwidth}
  \centering
  \includegraphics[width=.99\linewidth]{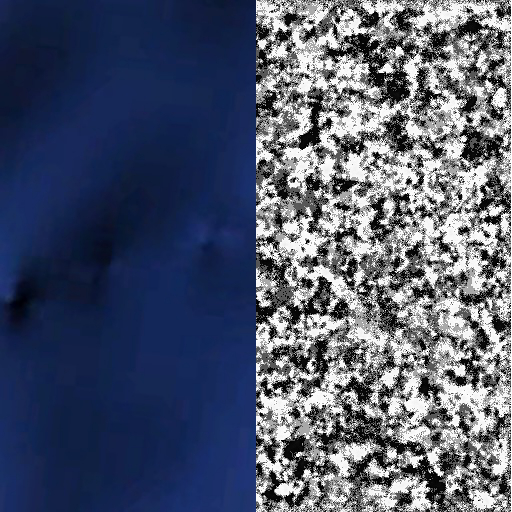}  
  \caption{IRCNN}
  \label{fig:sub_IRCNN_RICE}
\end{subfigure}
\begin{subfigure}{.16\textwidth}
  \centering
  \includegraphics[width=.99\linewidth]{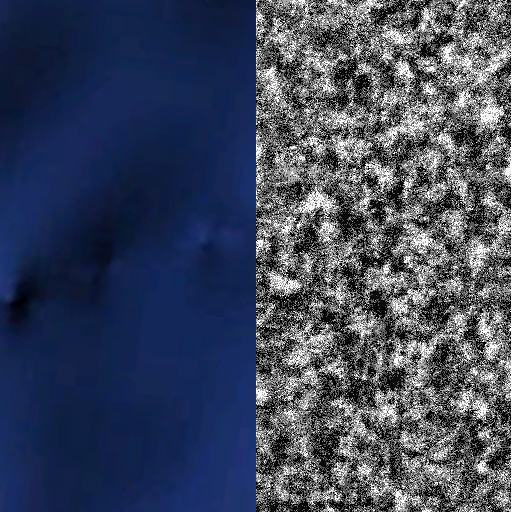}
  \caption{MemNet}
  \label{fig:sub_MemNet_RICE}
\end{subfigure}
\begin{subfigure}{.16\textwidth}
  \centering
  \includegraphics[width=.99\linewidth]{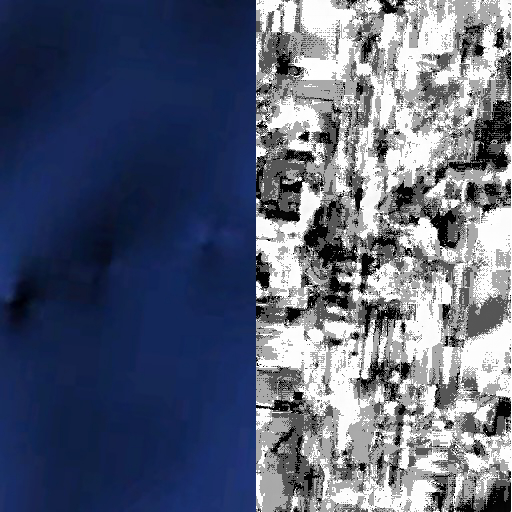}
  \caption{\ourmodel{} (Ours)}
  \label{fig:sub_FDN_RICE}
\end{subfigure}
\vspace{-2mm}\\

\caption{Visual results on RICE1 with $\sigma=70$. For (a), the left part is the clean image, and the right part is the noise. For~(b)-(f), the left part is the denoised image, and the right region reflects the difference between the denoised and GT images. Whiter pixels represent better denoising performance. The denoised image restored by \ourmodel{} is more closer to the ground truth.}
\label{fig:Visual_Rice}
\end{figure*}

\subsection{Remote Sensing Image Denoising}
\label{sec:Visual_Rice}

\textbf{Quantitative Results.} 
The results of denoising the RICE1 and RICE2~\cite{RICE} datasets with $\sigma$ =30, 50, 70 are reported in~\autoref{Tab:RICE}. \ourmodel{} improves by 1.13 dB-1.74 dB on RICE1 with different noise levels compared with the highest results from other competitive methods. On RICE2, \ourmodel{}outperforms other methods when the noise levels are large, \ie achieving an increase of 0.97 dB and 1.55 dB for $\sigma$ =50 and 70, respectively. These results demonstrate that \ourmodel{} is also capable of restoring images following complex distributions. 

\textbf{Qualitative Results.} 
The visual results are illustrated in~\autoref{fig:Visual_Rice}\footnote{Although in RGB images, blacker pixels represent smaller values, we change every pixel in the right regions by using 255 minus the value, and thus whiter regions are smaller. }. The remote sensing datasets are mainly composed of images with two types of regions: the texture regions such as mountains, and the smooth regions, for example, deserts. An example of the smooth region from RICE1 with $\sigma=70$ is taken. Our \ourmodel{} outperforms other methods significantly from the right regions of~\autoref{fig:sub_DnCNN_RICE1}-\ref{fig:sub_FDN_RICE}.
Thus, our distribution learning and disentanglement based denoising method \ie~\ourmodel{} has proven to be effective 
not only for category-specific data, but also for images following more complex distributions. 

\begin{table*}
\centering
\caption{Quantitative comparison on the real noisy SIDD dataset trained on SIDD medium dataset.} 
\resizebox{\textwidth}{!}{
\begin{tabular}{lrrrrrrrrrr}
\hline
\hline
Method & DnCNN~\cite{DnCNN} & TNRD~\cite{TNRD} & BM3D~\cite{BM3D} & CBDNet~\cite{Guo2019Cbdnet} & GradNet~\cite{liu2020gradnet} & AINDNet~\cite{AINDNet} & VDN~\cite{VDN} & \ourmodel{} (Ours)\\ 
\hline
PSNR (dB) & 23.66 & 24.73 & 25.65 & 33.28 & 38.34 & 39.08 & 39.26 & \textbf{39.31} \\
SSIM & 0.583 & 0.643 & 0.685 & 0.868 & 0.953 & 0.955 & 0.955 & \textbf{0.955} \\
Param (M) & 0.56 & -- & -- & 4.34 & 1.60 & 13.76 & 7.81  & 4.38 \\
Inference time (GFlops) & 73.32 & --  & -- & 80.76 &  213.06 & -- & 99.00 & 76.80 \\
\hline
\hline
\end{tabular}
}
\label{Tab:SIDD}
\end{table*}

\begin{figure*}[ht!]
\centering
\begin{subfigure}{.19\textwidth}
  \centering
  \includegraphics[width=.99\linewidth]{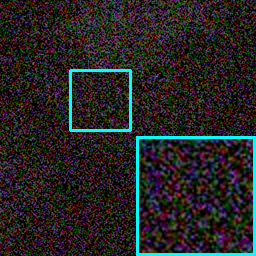}  
\end{subfigure}
\begin{subfigure}{.19\textwidth}
  \centering
  \includegraphics[width=.99\linewidth]{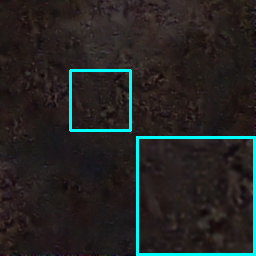}  
\end{subfigure}
\begin{subfigure}{.19\textwidth}
  \centering
  \includegraphics[width=.99\linewidth]{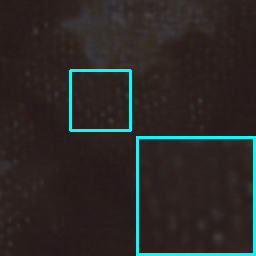}  
\end{subfigure}
\begin{subfigure}{.19\textwidth}
  \centering
  \includegraphics[width=.99\linewidth]{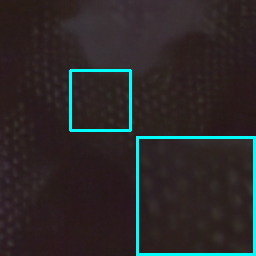}  
\end{subfigure}
\begin{subfigure}{.19\textwidth}
  \centering
  \includegraphics[width=.99\linewidth]{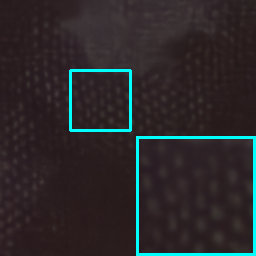}
\end{subfigure}
\vspace{1mm}\\

\begin{subfigure}{.19\textwidth}
  \centering
  \includegraphics[width=.99\linewidth]{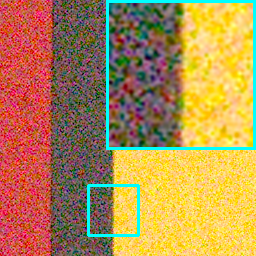}  
\end{subfigure}
\begin{subfigure}{.19\textwidth}
  \centering
  \includegraphics[width=.99\linewidth]{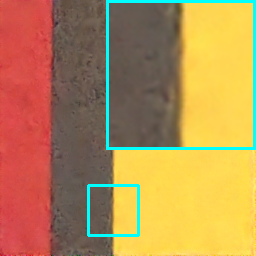}  
\end{subfigure}
\begin{subfigure}{.19\textwidth}
  \centering
  \includegraphics[width=.99\linewidth]{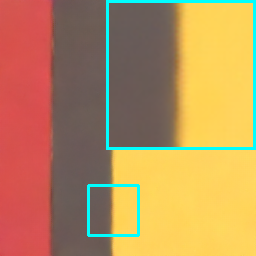}  
\end{subfigure}
\begin{subfigure}{.19\textwidth}
  \centering
  \includegraphics[width=.99\linewidth]{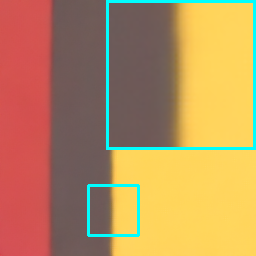}  
\end{subfigure}
\begin{subfigure}{.19\textwidth}
  \centering
  \includegraphics[width=.99\linewidth]{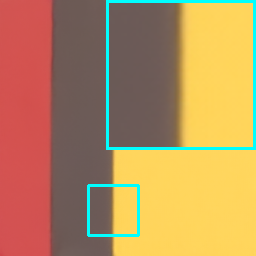}
\end{subfigure}
\vspace{1mm}\\

\begin{subfigure}{.19\textwidth}
  \centering
  \caption*{Noisy} 
\end{subfigure}
\begin{subfigure}{.19\textwidth}
  \centering
  \caption*{CBDNet} 
\end{subfigure}
\begin{subfigure}{.19\textwidth}
  \centering
  \caption*{GradNet} 
\end{subfigure}
\begin{subfigure}{.19\textwidth}
  \centering
  \caption*{VDN}   
\end{subfigure}
\begin{subfigure}{.19\textwidth}
  \centering
  \caption*{\ourmodel{} (Ours)} 
\end{subfigure}

\caption{The visual comparison on the SIDD dataset against state-of-the-art methods. In the first row, \ourmodel{} reconstructs the white dot patterns clearly in a dark environment without smoothing and artifacts. In the second row, \ourmodel{} preserves more crisp edges. }
\label{fig:sidd}
\end{figure*}


\vspace{-5mm}
\subsection{Real Image Denoising} 
\textbf{Quantitative Results.}
The performance comparison on the test set of the real noise dataset SIDD~\cite{SIDD_2018_CVPR} is listed in ~\autoref{Tab:SIDD}. 
We achieve a new state-of-the-art denoising accuracy comparing with other methods. In addition, our model size (4.38M) is much smaller than the competitive AINDNet~\cite{AINDNet} (13.76M) and VDN~\cite{VDN} (7.81M), illustrating that \ourmodel{} is suitable to be deployed on small edge devices. We also report the inference time (in GigaFlops) of one $256 \times 256$ image for each method. \ourmodel{} is much faster than VDN~\cite{VDN}.

\textbf{Qualitative Results.}
To further present the effectiveness of \ourmodel{} against other state-of-the-art methods, we show the visual results of denoised images in~\autoref{fig:sidd}.
\ourmodel{} restores accurate textures and well-shaped edges, while other methods blur details and introduce artifacts. 
This indicates that \ourmodel{} is also superior in removing real-world noise.

\section{Conclusion}
The widely used image denoising CNNs are discriminative models, learning the mapping between noisy images and their clean counterparts via learning features of images. However, these methods may overlook the underlying distribution of the clean ground truth, resulting in downgraded visual results with blurry regions or artifacts.
This paper provides a new perspective to understand image denoising as a distribution disentangling task. Since the distribution of noisy images can be treated as a joint distribution of clean images and noise, the denoised images can be obtained via the clean images' latent representations. A distribution learning-based denoising framework is proposed in this paper.
We also present a noval denoising network, FDN, based on normalizing flows without adding any assumptions on clean images and noise distributions. \ourmodel{} learns the distribution instead of features from noisy images, which is different from the previous feature learning based networks. A distribution disentanglement method for denoising is introduced as well. Experimental results verify the effectiveness of \ourmodel{} on both category-specific and remote sensing image denoising with synthetic AWGN. Moreover, \ourmodel{} shows its superiority in real image denoising with fewer parameters and lower running time. 
In conclusion, this paper presents a new potential direction to optimize image denoising methods in the future.





\ifCLASSOPTIONcaptionsoff
  \newpage
\fi









\end{document}